\documentclass[lettersize,journal]{IEEEtran}
\usepackage{amsmath,amsfonts}
\usepackage{algorithmic}
\usepackage{algorithm}
\usepackage{array}
\usepackage[caption=false,font=normalsize,labelfont=sf,textfont=sf]{subfig}
\usepackage{textcomp}
\usepackage{stfloats}
\usepackage{url}
\usepackage{verbatim}
\usepackage{graphicx}
\usepackage{cite}
\usepackage{pifont}

\newcommand{\xmark}{\ding{55}}
\usepackage[table]{xcolor}
\usepackage{multirow}
\usepackage{booktabs}
\usepackage{epsfig}
\usepackage{hyperref}
\usepackage{bm}
\usepackage[normalem]{ulem}
\definecolor{lightblue}{rgb}{0.85, 0.92, 0.98}
\usepackage{orcidlink}

\begin{document}

\title{Reconstructing Building Height from Spaceborne TomoSAR\\Point Clouds Using a Dual-Topology Network}

\author{Zhaiyu~Chen~\orcidlink{0000-0001-7084-0994}, 
        Yuanyuan~Wang~\orcidlink{0000-0002-0586-9413},~\IEEEmembership{Member,~IEEE,}
        Yilei~Shi~\orcidlink{0000-0003-1907-8214},~\IEEEmembership{Member,~IEEE,}\\
        and~Xiao~Xiang~Zhu~\orcidlink{0000-0001-5530-3613},~\IEEEmembership{Fellow,~IEEE}\\
\thanks{The work is jointly supported by the TUM Georg Nemetschek Institute under the AI4TWINNING project, by the European Commission through the project ``MultiMiner: Multi-source and multi-scale Earth observation and novel machine learning methods for mineral exploration and mine site monitoring'' under the Horizon 2020 Research and Innovation program (Grant Agreement No. 101091374) and by the Munich Center for Machine Learning. \textit{(Corresponding author: Xiao Xiang Zhu.)}}
\thanks{Z. Chen, Y. Wang, and X. X. Zhu are with the Chair of Data Science in Earth Observation, Technical University of Munich (TUM), 80333 Munich, Germany (e-mail: zhaiyu.chen@tum.de; y.wang@tum.de; xiaoxiang.zhu@tum.de). Z. Chen and X. X. Zhu are also with the Munich Center for Machine Learning (MCML). Y. Shi is with the School of Engineering and Design, Technical University of Munich (TUM), 80333 Munich, Germany (e-mail: yilei.shi@tum.de).}
\thanks{This work has been submitted to the IEEE for possible publication. Copyright may be transferred without notice, after which this version may no longer be accessible.}
}

\markboth{Accepted for publication in IEEE Transactions on Geoscience and Remote Sensing}%
{Z. Chen \MakeLowercase{\textit{et al.}}: TomoSAR2Height}

\maketitle

\begin{abstract}
Reliable building height estimation is essential for various urban applications. Spaceborne SAR tomography (TomoSAR) provides weather-independent, side-looking observations that capture facade-level structure, offering a promising alternative to conventional optical methods. However, TomoSAR point clouds often suffer from noise, anisotropic point distributions, and data voids on incoherent surfaces, all of which hinder accurate height reconstruction. To address these challenges, we introduce a learning-based framework for converting raw TomoSAR points into high-resolution building height maps.
Our dual-topology network alternates between a point branch that models irregular scatterer features and a grid branch that enforces spatial consistency. By jointly processing these representations, the network denoises the input points and inpaints missing regions to produce continuous height estimates. To our knowledge, this is the first proof of concept for large-scale urban height mapping directly from TomoSAR point clouds. Extensive experiments on data from Munich and Berlin validate the effectiveness of our approach. Moreover, we demonstrate that our framework can be extended to incorporate optical satellite imagery, further enhancing reconstruction quality. The source code is available at \url{https://github.com/zhu-xlab/tomosar2height}.
\end{abstract}



\section{Introduction}

Large-scale 3D modeling of the built environment is essential for diverse applications such as urban planning, disaster management, and environmental monitoring. A critical aspect of this modeling is the reliable estimation of building heights. Traditionally, airborne LiDAR scanning and photogrammetry have been employed to obtain high-quality height data. However, these techniques suffer from limited scalability. LiDAR surveys incur high costs, and photogrammetric methods require extensive collections of cloud-free, high-resolution optical images. Although recent advances in computer vision have enabled height estimation from single images~\cite{chen_htc-dc_2023, cao2024deep, yadav2025high}, these monocular approaches remain hampered by their dependence on clear-sky conditions and by the strong inductive biases required to resolve depth ambiguities.

Spaceborne synthetic aperture radar (SAR) provides a complementary data source for large-scale 3D reconstruction, thanks to its all-weather imaging capability and its ability to capture 3D structure. In particular, multi-baseline SAR tomography (TomoSAR) extends conventional interferometry by reconstructing fully three-dimensional reflectivity profiles, thereby separating overlapping scatterers within a single ground resolution cell~\cite{fornaro_four-dimensional_2009, zhu_tomographic_2010}. Leveraging meter-resolution SAR imagery from modern satellites (\textit{e.g.}, TerraSAR-X and TanDEM-X), TomoSAR can produce consistent large-scale point clouds of urban areas~\cite{shi_towards_2018, zhu_towards_2018}. These SAR-derived point clouds offer distinct geometric insights, most notably by capturing building facades through the side-looking acquisition geometry, which are often missed by nadir-view sensors. In addition, TomoSAR point clouds may provide high geolocation accuracy, especially with advanced calibration~\cite{zhu2016}.

\begin{figure}[ht]
  \centering    \centerline{\epsfig{figure=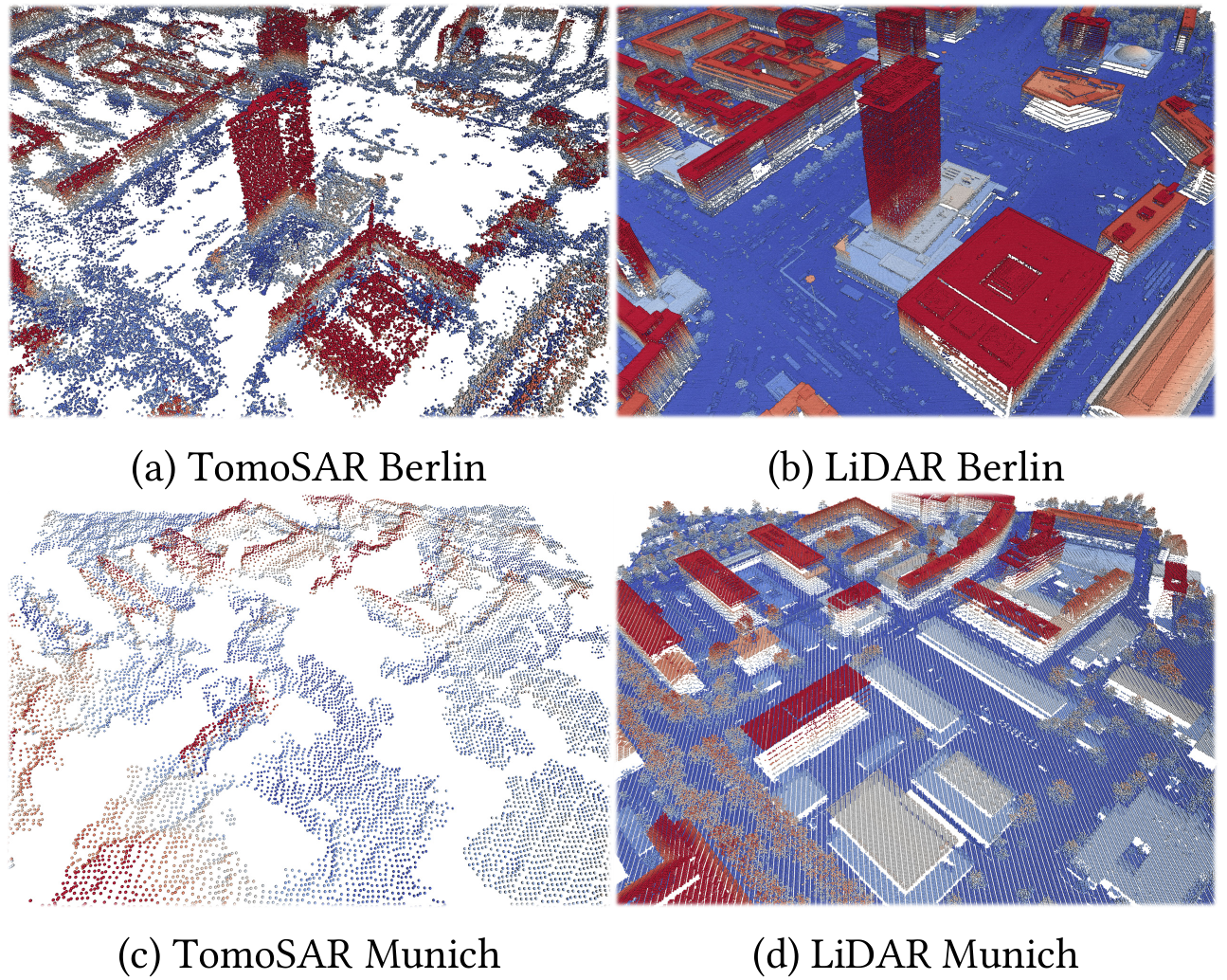,width=0.95\linewidth}}
\caption{Comparison of spaceborne TomoSAR point clouds and airborne LiDAR data over the same areas in Berlin (a, b) and Munich (c, d). The two point clouds were reconstructed from SAR image stacks of different sizes and spatial resolutions, resulting in different quality. While TomoSAR point clouds provide extensive coverage, they exhibit higher noise and a more heterogeneous point distribution, requiring specialized processing techniques. Points are color-coded by height.}
\label{fig:difference}
\end{figure}

Despite these advantages, TomoSAR point clouds pose significant challenges for building height reconstruction. The inherent imaging process and side-looking geometry often lead to data that are sparse and noisy, with uneven point densities and gaps, particularly over less coherent surfaces such as building roofs. These challenges become more pronounced when using lower-resolution SAR imagery or a limited number of acquisitions. As shown in \autoref{fig:difference}, TomoSAR point clouds deliver wide-area coverage and rich facade detail but typically exhibit lower point density and higher noise than airborne LiDAR. These challenges undermine conventional height mapping techniques such as spatial interpolation or geometric fitting. Accurately reconstructing building heights from TomoSAR therefore requires specialized methods capable of denoising sparse, anisotropically sampled points and inpainting the data voids.

To address these challenges, we propose a neural network designed to reconstruct building height maps directly from spaceborne TomoSAR point clouds. Since the horizontal geolocation of TomoSAR points is typically more reliable than their vertical elevation, we adopt a dual-topology design that pairs a point-based branch with an $x$--$y$ grid branch, preserving structural detail from irregular points while using the grid to regularize noisy heights and enforce spatial consistency. This design enables high-resolution height mapping without requiring an external digital terrain model (DTM) at inference. Moreover, the dual representation strengthens denoising and inpainting, mitigating issues caused by noise, anisotropic point distributions, and data voids. The pipeline is inherently extensible, and we demonstrate that incorporating optical satellite imagery provides complementary information and further improves reconstruction quality.

Experimental validation over urban areas in Munich and Berlin demonstrates the effectiveness of our method under varying data acquisition conditions. These results underscore the potential of our approach as a proof of concept towards operational, large-scale building height mapping.

Our primary contributions are summarized as follows: 
\begin{itemize}
\item We introduce a learning-based framework for large-scale reconstruction of building height maps from spaceborne TomoSAR point clouds.
\item We present a dual-topology neural network that alternates between a point topology for modeling irregular scatterer features and a grid topology for enforcing spatial consistency, enabling effective denoising and inpainting.
\item We demonstrate the extensibility of our framework by integrating optical imagery, which further improves height reconstruction and reinforces its potential for large-scale urban mapping.
\end{itemize}


\section{Related work}

Building modeling typically depends on high-precision 3D data, such as point clouds obtained from airborne LiDAR and photogrammetry~\cite{chen2024polygnn, stucker2022implicity}. However, acquiring such data at scale is costly and often infeasible for many areas. For many urban applications, having only building height data would suffice. Over the past years, alternative methods such as height estimation from single images and from SAR interferometry have become available, because of the development of deep learning techniques and the availability of high-resolution repeat-pass spaceborne SAR images.

\subsection{Single-image height estimation}
To make height information more accessible, researchers have developed methods using monocular optical images. For instance, Mou and Zhu proposed a model that uses a fully convolutional network to deduce a digital surface model from a single satellite image~\cite{mou_im2height_2018}. Subsequent works explored generative adversarial networks~\cite{ghamisi_img2dsm_2018, paoletti2020u}, multi-task learning~\cite{srivastava2017joint, xiong2022disentangled}, and hybrid regression~\cite{chen_htc-dc_2023} to improve monocular height estimation. These monocular optical image methods hold promise but require unobstructed and preferably high-resolution images. Although feasible in several applications~\cite{li2020developing, huang2022estimating, wu2023first, cao2024deep}, producing a timely global high-resolution height map solely from optical imagery is challenging due to inconsistent image quality and frequent cloud cover.

SAR provides an attractive alternative for building height estimation, thanks to its day-and-night imaging and general all-weather availability. Consequently, a number of studies aim to estimate urban building heights directly from SAR images. For example, Recla and Schmitt introduced a deep network that learns to predict a height map from a single very-high-resolution SAR image~\cite{recla2024sar2height,recla2022deep}, and Sun \textit{et al.} employed bounding-box regression to retrieve building heights when building footprints are known~\cite{sun2022large}. Fusion approaches that combine SAR and optical data have also been explored~\cite{yadav2025high}. Although these methods enable rapid coverage, they inherently lack explicit 3D geometry, which can lead to ambiguities.

While monocular optical and SAR single‐image approaches have lowered the barrier to urban height mapping, their dependence on 2D observations limits robustness in complex urban scenes. In contrast, interferometric SAR (InSAR) techniques leverage multiple viewing angles to disentangle overlapping scatterers and directly recover 3D structures. In particular, multi‐pass TomoSAR has emerged as a promising solution for detailed urban reconstruction.

\subsection{TomoSAR for urban reconstruction}
InSAR can generate large-scale digital elevation models, but it struggles in dense urban environments due to layover, where ground and building signals overlap, making it unable to separate multiple scatterers within a single resolution cell~\cite{krieger2007tandem}. Multi-pass TomoSAR addresses this limitation by using a stack of SAR images at slightly different viewing angles to reconstruct the 3D distribution of scatterers. Widely recognized as a powerful method for urban area reconstruction, TomoSAR can resolve multiple reflective targets per resolution cell and produce detailed 3D point clouds~\cite{fornaro2005three,fornaro_four-dimensional_2009,zhu2010very}. Early TomoSAR research introduced model-based inversion techniques to recover reflectivity profiles from spaceborne data. For instance, Zhu and Bamler demonstrated high-resolution TomoSAR for urban areas using TerraSAR-X data~\cite{zhu2010very}, while Fornaro \textit{et al.} developed multi-pass focusing methods for estimating single and double scatterer heights~\cite{fornaro_four-dimensional_2009}. Although these model-based approaches can achieve precise reconstructions, they often struggle to separate closely spaced scatterers and typically require dozens of images. To tackle closely spaced scatterers, compressive sensing techniques were introduced for TomoSAR inversion~\cite{zhu_tomographic_2010}. In particular, Zhu and Bamler’s L1-norm regularization approach yields super-resolved elevation estimates, improving reconstruction accuracy and scatterer separation~\cite{zhu_tomographic_2010}. While TomoSAR benefits from SAR's general all-weather capability, it should be noted that, as a multi-baseline multi-temporal InSAR technique, it is sensitive to small phase changes, particularly atmospheric delays, which can be pronounced in tropical regions. Fortunately, recent studies have shown that reliable reconstructions are possible even with as few as 3--5 interferograms~\cite{shi2020sar}, significantly lowering the TomoSAR data demand. More recently, Qian \textit{et al.} introduced learning-based TomoSAR inversion frameworks that mimic or accelerate compressive sensing, boosting processing efficiency without notable accuracy loss~\cite{qian_gamma-net_2022,qian_basis_2022,qian_hyperlista-abt_2024}. Thanks to these developments and the growing availability of high-resolution SAR imagery, extensive TomoSAR point clouds for major cities worldwide have become feasible, capturing 3D structural information at an unprecedented scale~\cite{wang_fusing_2017,jiao2020urban}.

Given a TomoSAR point cloud of an urban area, one can extract a wealth of building-related information beyond individual scatterers. For instance, Yang \textit{et al.} applied Pol-TomoSAR to estimate heights in forested regions~\cite{yang2023deep}, while Armeshi \textit{et al.} employed a TomoSAR-based regularization method for detecting height changes in urban settings~\cite{armeshi2024tomosar}. However, neither study generated nor leveraged 3D TomoSAR point clouds for their analyses. By contrast, Shahzad and Zhu demonstrated the automatic reconstruction of facades and 3D building shapes from spaceborne TomoSAR data, confirming that building geometries can be inferred from such point clouds~\cite{shahzad2014robust, shahzad2015automatic}. Ley \textit{et al.}~\cite{ley2018regularization} proposed a convex optimization approach to denoise TomoSAR point clouds and fill gaps in derived height maps. Despite these advances, to our knowledge, no learning-based method has been developed to directly generate continuous building height maps from 3D TomoSAR point clouds.

TomoSAR point clouds are obtained by coherently combining multiple SAR acquisitions along the elevation axis, integrating all echoes within a vertical resolution cell into a single response. This acquisition geometry leads to strongly anisotropic noise: elevation ($z$) errors are typically about one order of magnitude larger than those in $x$ and $y$~\cite{zhu2011super}. With advanced point-wise analysis and atmospheric correction, absolute planimetric accuracy can reach the centimeter level~\cite{eineder2011}, but the effective height accuracy is much lower because the position of the elevation peak depends on the number and spatial distribution of acquisitions and the scatterer SNR. For TerraSAR-X stacks, vertical uncertainties are typically on the order of 1--20\,m. As a result, TomoSAR point clouds are substantially noisier, sparser, and less evenly distributed than typical photogrammetric or LiDAR point clouds, especially over low-coherence surfaces. Simple interpolation is therefore unreliable, and purely local processing is insufficient: the model must capture global point-cloud patterns and exploit spatial context to denoise, fill voids, and suppress spurious scatterers. Since horizontal location is much more reliable than vertical height, it is natural to use a dual-topology design that pairs a point-based branch with a grid-based branch in the $x$--$y$ map plane, so that the grid regularizes noisy elevations and enforces spatial consistency. Peng \textit{et al.} proposed fusing irregular 3D points into a continuous occupancy grid using convolutional encoders~\cite{peng2020eccv}, while Wang \textit{et al.} introduced an alternating strategy to iteratively refine both the grid and point representations~\cite{wang2023alto}. Our model extends these ideas with separate point and grid branches but tailored to 2.5D height mapping: we project all scatterers onto a single nadir plane instead of the original tri-plane scheme, and replace the occupancy decoder with a refinement module that directly outputs continuous height values. The point- and grid-based feature transformations are arranged in a U-Net cascade~\cite{ronneberger2015u}, so that at each stage both streams evolve alternately to denoise, fill voids, and enforce spatial consistency.

\section{Methodology}

\subsection{Problem definition}

Our objective is to derive building height maps from TomoSAR point clouds and represent the results as normalized digital surface models (nDSM). This requires learning a function that maps the input points into grid-based height values while coping with large data gaps and anisotropic noise. To this end, we design a neural network that predicts pixel-wise heights directly.
Let $\mathbf{P} = \{\mathbf{p}_i\}_{i=1}^N$ denote the set of points from TomoSAR, with each point $\mathbf{p}_i = (x_i, y_i, z_i)$ featuring spatial coordinates $(x_i, y_i)$ and an elevation $z_i$. We seek a mapping $f$ such that
\begin{equation} 
\hat{\mathbf{H}} = f(\mathbf{P}), 
\end{equation}
where $\hat{\mathbf{H}} = \{ \hat{h}_j \}_{j=1}^{M}$ denotes the estimated height at location $(x_j, y_j)$ on the regular grid, with $M$ being the number of grid cells. Note that the predictions $\hat{h}_j$ are defined on grid cells rather than on individual points.

\begin{figure}[ht]
  \centering
  \includegraphics[width=0.99\linewidth]{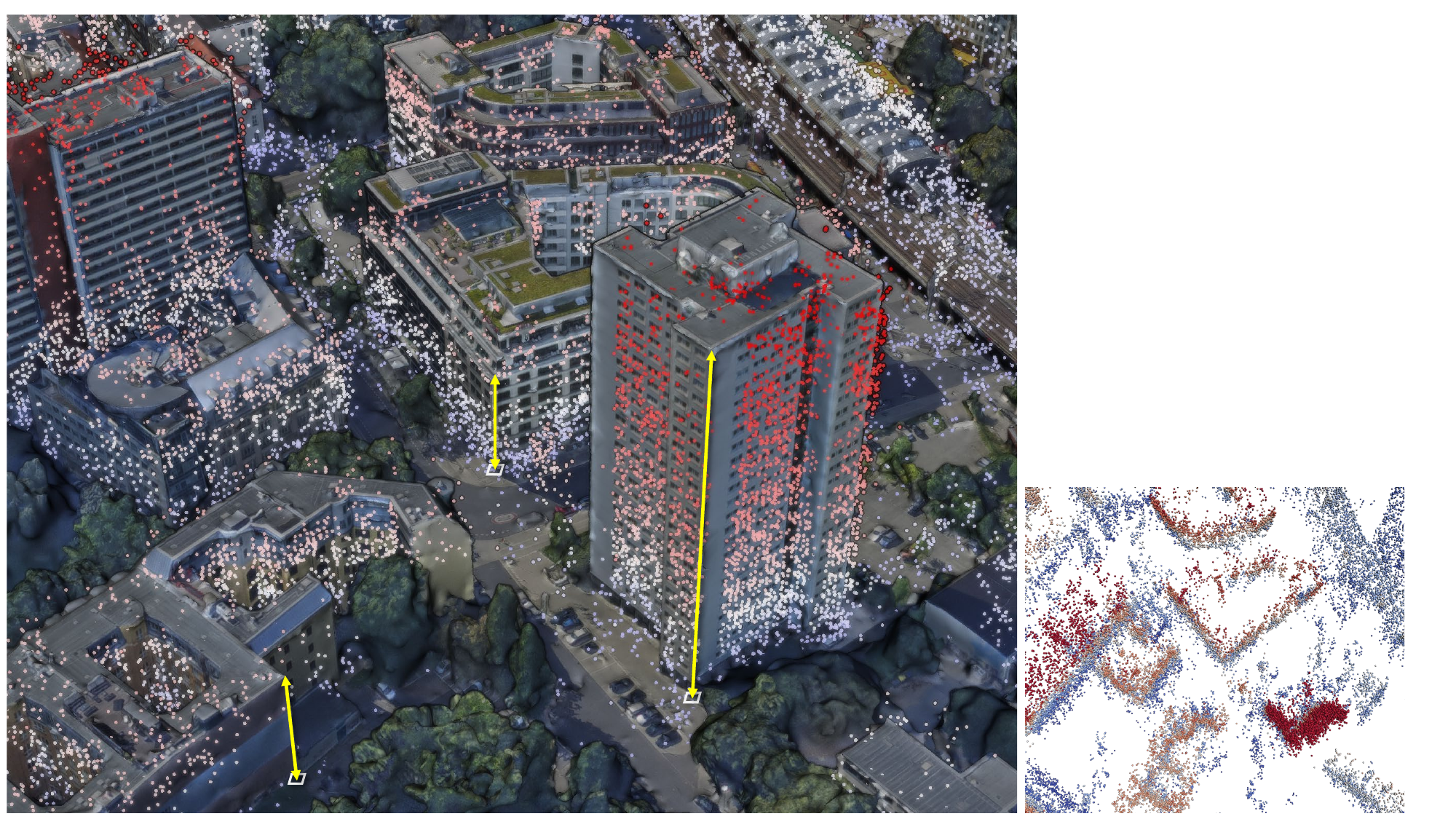}
  \caption{Building heights can be inferred by holistically analyzing facade and neighborhood point patterns in a TomoSAR point cloud. At this Berlin site, three pixel-wise height values are highlighted. The right panel shows a top-down view. For visualization, points are color-coded by height and slightly offset from the facades.}
  \label{fig:facade}
\end{figure}

The primary challenges associated with TomoSAR point clouds, particularly those derived from the stripmap mode with a limited number of acquisitions, are high noise levels and anisotropic sampling. The noise degrades point localization, making it necessary to exploit broader spatial context. Anisotropic sampling further complicates reconstruction: some areas lack height information entirely, while others may contain multiple candidate elevations due to the slant-range imaging geometry, as shown in \autoref{fig:difference}.

Straightforward solutions, such as estimating height at each $(x_j, y_j)$ by subtracting terrain elevation from the local maximum, are highly sensitive to noise. Spatial interpolation is also unreliable under anisotropic sampling (see \autoref{fig:motivation}). Tackling these issues necessitates a holistic understanding of the point patterns. We argue that, by analyzing the facade and neighborhood point distributions, building heights can be inferred without relying on precise point positions or a reference DTM at inference, as illustrated in \autoref{fig:facade}. These challenges motivate a robust model capable of both denoising the input points and inpainting missing values to reconstruct a reliable building height map $\hat{\mathbf{H}}$.

\subsection{Proposed solution}

\begin{figure*}[ht]
  \centering
\centerline{\epsfig{figure=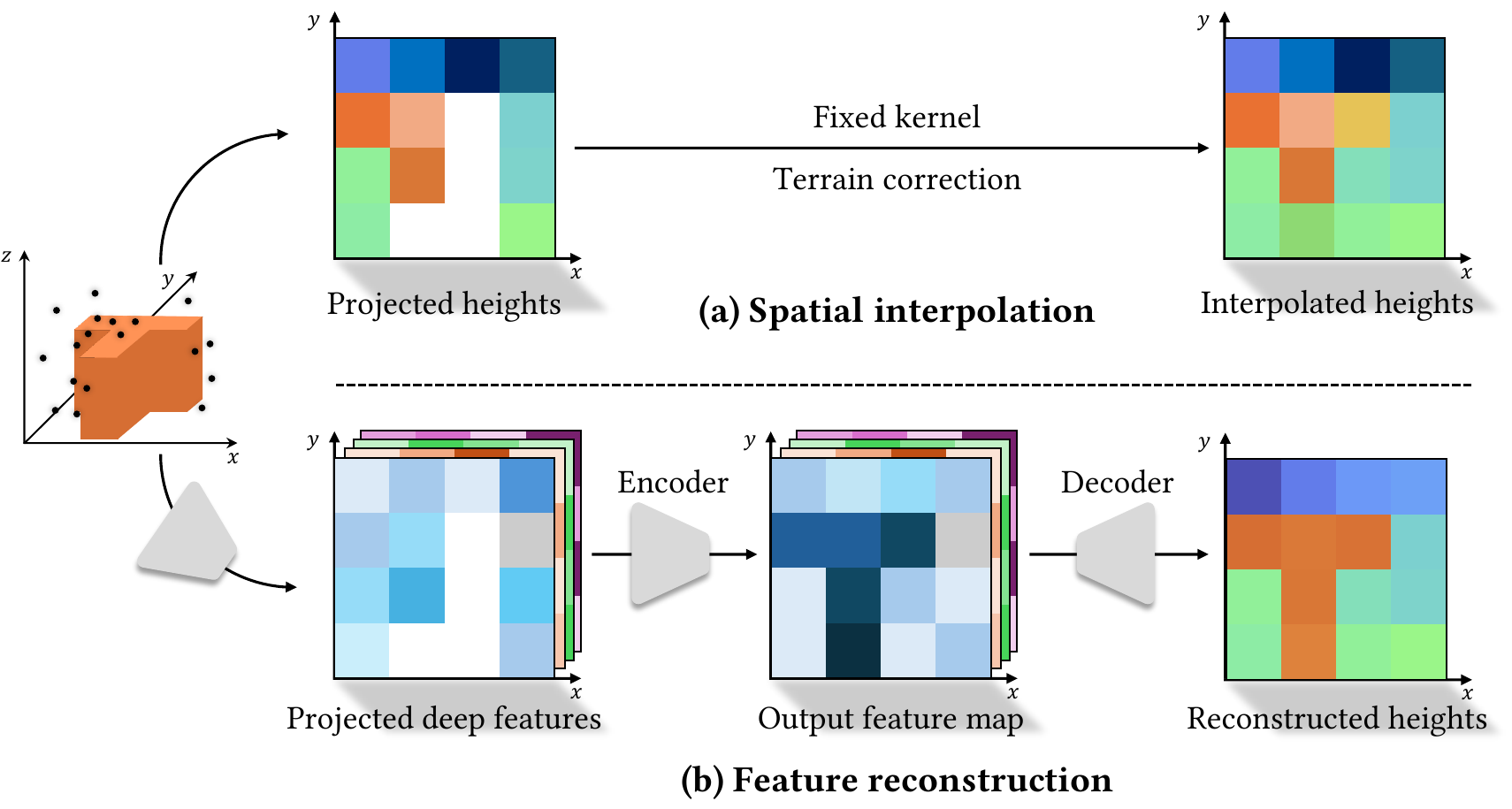,width=0.80\linewidth}}
\caption{Comparison of traditional spatial interpolation with our feature-based height reconstruction. Spatial interpolation (a) attends to explicit height values, making it vulnerable to missing or noisy point distributions and necessitating terrain correction. In contrast, our approach (b) leverages inductive biases to refine and complete the deep features encoded from input points, resulting in more robust and accurate height reconstruction.}
\label{fig:motivation}
\end{figure*}

We address the challenges using a data-driven approach. Rather than directly interpolating heights in the spatial domain, we elevate the problem to a deep feature space, allowing a deep neural network to identify comprehensive point patterns, denoise observations, and inpaint missing regions. This approach leverages the inductive biases inherent in modern neural networks. The motivation is illustrated in \autoref{fig:motivation}.

\begin{figure*}[ht]
  \centering
\centerline{\epsfig{figure=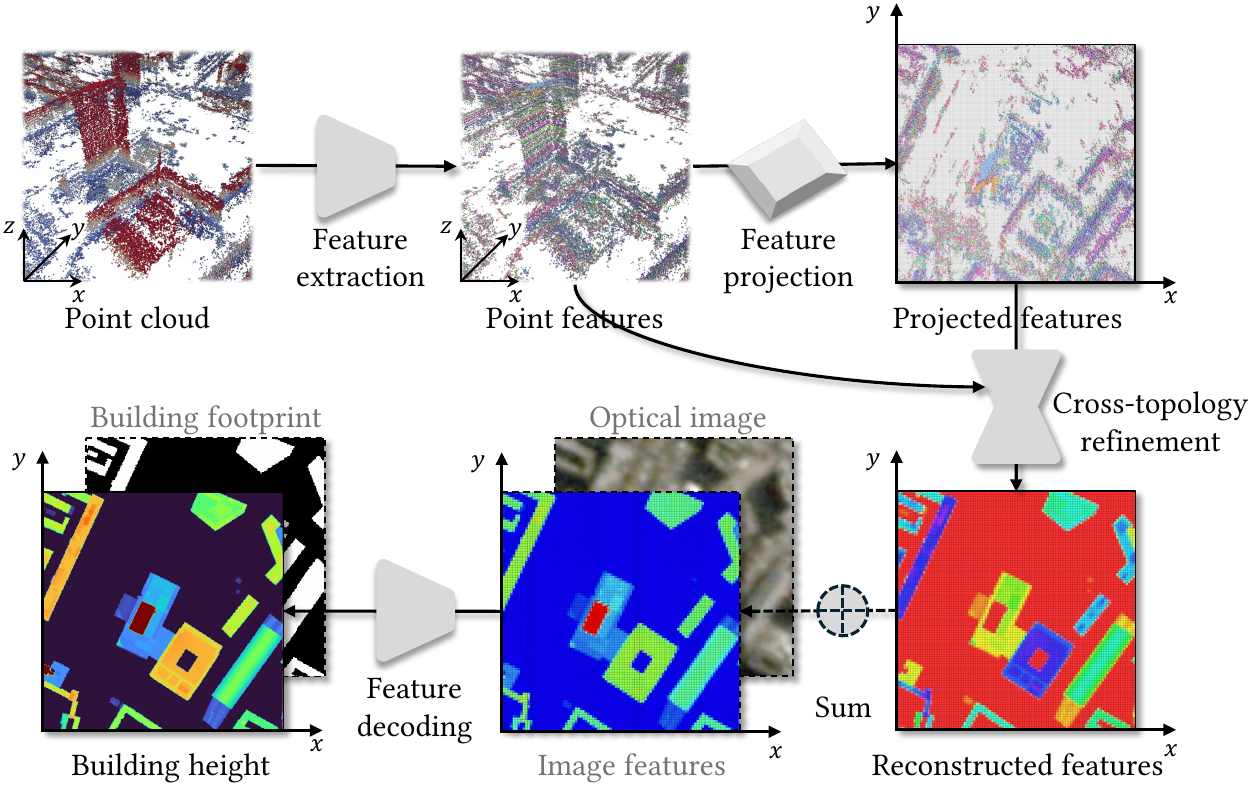,width=0.72\linewidth}}
\caption{Overview of the proposed workflow. Starting from a TomoSAR point cloud, we first extract point-wise features and locally pool them. The features are then projected onto a 2D grid to form a horizontal feature plane, where cross-topology refinement iteratively improves the representation by exchanging information between points and grid cells. Finally, a grid-based decoder predicts building heights from the refined grid features, producing a coherent height map. Optional optical image features and building footprint supervision are indicated by dashed outlines.}
\label{fig:architecture}
\end{figure*}

To effectively handle sparse and noisy input points, our network comprises three primary components: (1) a point-to-grid encoder that extracts point features and aggregates them onto a grid; (2) a dual-topology refinement module that alternates between point and grid representations to progressively enhance the features; and (3) a grid feature decoder that produces the height map, optionally together with auxiliary outputs. \autoref{fig:architecture} provides an overview of the architecture and the data flow through these stages.

\subsubsection{Point-to-grid encoder}

We extract features from the input 3D points and map these features onto a 2D grid according to their spatial coordinates. This encoding transforms unorganized points into a structured representation.

\paragraph{Feature extraction}

\begin{figure}[htb]
  \centering
\centerline{\epsfig{figure=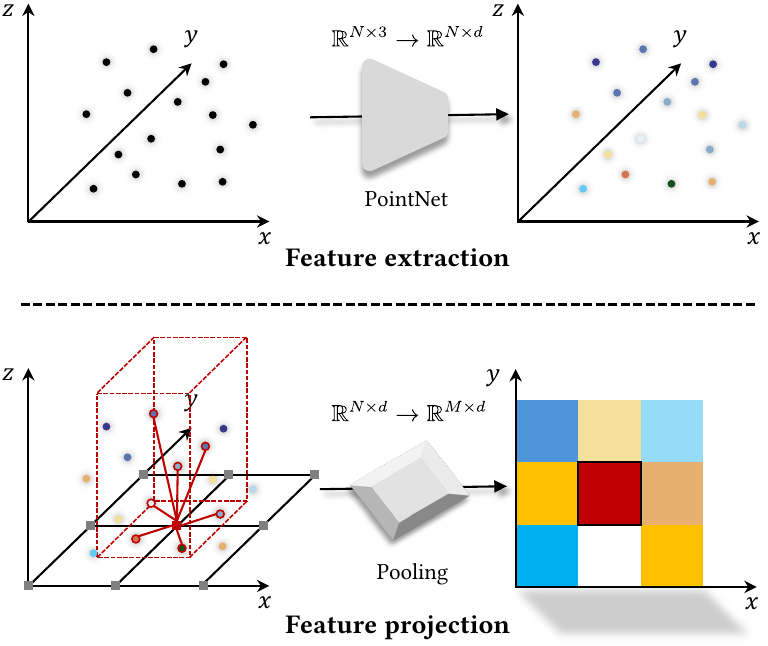,width=0.90\linewidth}}
\caption{Feature encoding process. \textit{Top:} Given a TomoSAR point cloud \((x,y,z)\) with \(N\) points, we first extract point-wise features \(\mathbf{Z} \in \mathbb{R}^{N \times d}\) using a PointNet. \textit{Bottom:} We then project these point features onto a 2D grid by average-pooling the features of all points that fall within each grid cell, producing grid features \(\mathbf{G} \in \mathbb{R}^{M \times d}\).}
\label{fig:encoding}
\end{figure}

To encode the points $\mathbf{P}$ into latent features that capture structural information beyond individual points, we employ an encoder network $f_e: \mathbb{R}^{N\times 3} \rightarrow \mathbb{R}^{N\times d}$ that exploits their spatial relationships and produces a set of $d$-dimensional features $\mathbf{Z} = \{\mathbf{z}_i\}_{i=1}^N$:
\begin{equation}
    \mathbf{Z} = f_e(\mathbf{P}),
    \label{eq:point_feature}
\end{equation}
where $f_e$ is implemented using a stack of PointNet layers~\cite{qi2017pointnet} with local pooling onto the grid, as illustrated in \autoref{fig:encoding}.

\paragraph{Feature projection}

The point features $\mathbf{Z}$ are then projected onto horizontal 2D grid features $\mathbf{G} = \{ \mathbf{g}_j \}_{j=1}^{M}$, where each $\mathbf{g}_{j}$ remains a $d$-dimensional vector. This projection is immediately followed by aggregating the projected features that fall onto the same cell, with average pooling. The chained process, represented by $\mathcal{P}roj$, is expressed as follows:
\begin{equation}
\mathcal{P}roj: \{\mathbf{z}_i\}_{i=1}^N \;\mapsto\; \{\mathbf{g}_{j}\}_{j=1}^{M},
\end{equation}
where $\text{cell}(j)$ denotes the spatial region covered by grid cell $j$ on the horizontal plane. The aggregated feature vector \(\mathbf{g}_j\) is given by
\begin{equation}
\mathbf{g}_{j} = \frac{1}{|\mathcal{I}(j)|} \sum_{\substack{k \in \mathcal{I}(j)}} \mathbf{z}_k,
\label{eq:grid_feature}
\end{equation}
where \(\mathcal{I}(j) = \{ k \mid (x_k,y_k) \in \text{cell}(j) \}\). \autoref{fig:encoding} (bottom) illustrates this projection.

\subsubsection{Cross-topology refinement}

\begin{figure}[ht]
  \centering
\centerline{\epsfig{figure=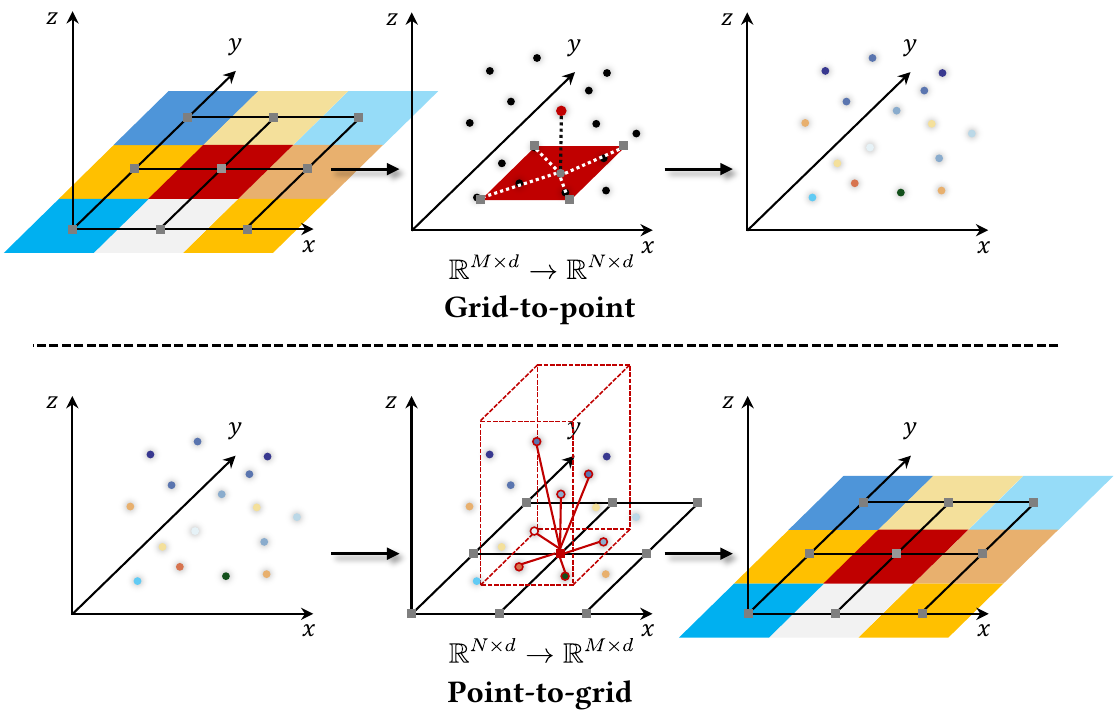,width=0.99\linewidth}}
\caption{Cross-topology feature refinement process. \textit{Top:} In the grid-to-point step, grid features \(\mathbf{G}^{(l)} \in \mathbb{R}^{M \times d}\) are interpolated back to the point domain \(\mathbf{Z}^{(l+1)} \in \mathbb{R}^{N \times d}\). \textit{Bottom:} In the point-to-grid step, the refined point features \(\mathbf{Z}^{(l+1)}\) are projected onto the grid, aggregating point-level information within each cell through pooling. This yields updated grid features \(\mathbf{G}^{(l+1)}\).}\label{fig:refinement}
\end{figure}

The initial grid features produced by the point-to-grid encoder are coarse and partially unreliable due to noise in the input points. Moreover, anisotropic sampling leaves many grid cells empty, whose features are therefore padded with zeros to maintain consistent feature dimensionality. To address these issues, we iteratively refine the grid features by alternating between the point-based representation $\mathbf{Z}^{(l)}$ and the grid-based representation $\mathbf{G}^{(l)}$, where $l \in \{0,\ldots,L\}$ denotes the iteration index. \autoref{fig:refinement} illustrates this cross-topology refinement. Here, \(\mathbf{Z}^{(0)}\) and \(\mathbf{G}^{(0)}\) are obtained from \autoref{eq:point_feature} and \autoref{eq:grid_feature}, respectively.

\paragraph{Grid-to-point transformation}

At iteration \(l\), the grid features \(\mathbf{G}^{(l)}\) are mapped back to point features \(\mathbf{Z}^{(l+1)}\), where each point \(\mathbf{p}_i \in \mathbf{P} \) is projected onto the 2D grid, and its feature \(\mathbf{z}_i^{(l+1)}\) is obtained via bilinear interpolation from the grid features. Let \(\mathcal{N}_i\) denote the four grid cells surrounding the projected location \((x_i,y_i)\), and let \(\alpha_{j}\) be the corresponding interpolation weights. Then,
\begin{equation}
\mathbf{z}_i^{(l+1)} = \sum_{j \in \mathcal{N}_i} \alpha_{j} \cdot \mathbf{g}^{(l)}_{j},
\end{equation}
where \(\mathbf{g}^{(l)}_j\) is the grid feature for the \(j\)-th cell. Collecting all point features gives \(\mathbf{Z}^{(l+1)} = \{\mathbf{z}_i^{(l+1)}\}_{i=1}^N\).

\paragraph{Point-to-grid transformation}

Given the updated point features \(\mathbf{Z}^{(l+1)}\), we first process them through an MLP to achieve finer granularity:
\begin{equation}
\tilde{\mathbf{z}}_i^{(l+1)} = \text{MLP}\left(\mathbf{z}_i^{(l+1)}\right).
\end{equation}
We then project the refined point features back onto the grid and aggregate the features with average pooling:
\begin{equation}
\mathbf{g}_j^{(l+1)} = \frac{1}{|\mathcal{I}(j)|} \sum_{k \in \mathcal{I}(j)} \tilde{\mathbf{z}}_k^{(l+1)}.
\end{equation}
This yields updated grid features \(\mathbf{G}^{(l+1)} = \{\mathbf{g}_j^{(l+1)}\}_{j=1}^M\).

To preserve detailed features, skip connections are integrated across successive transformations. Following the alternating design of Wang \textit{et al.} \cite{wang2023alto}, we arrange these transformations in a U-Net~\cite{ronneberger2015u} style. As refinement proceeds, \(\mathbf{G}^{(l)}\) and \(\mathbf{Z}^{(l)}\) evolve jointly, refining both the grid features and the point features.

\subsubsection{Grid feature decoder}

The final refined grid features \(\mathbf{G}^{(L)} \in \mathbb{R}^{M \times d}\) are used to reconstruct the building height map $\hat{\mathbf{H}}$. In addition, we introduce an auxiliary branch that predicts a building footprint $\hat{\mathbf{A}} = \{ \hat{a}_j \}_{j=1}^{M}$, providing extra supervision that improves robustness to noise.

\begin{figure*}[ht]
  \centering
\centerline{\epsfig{figure=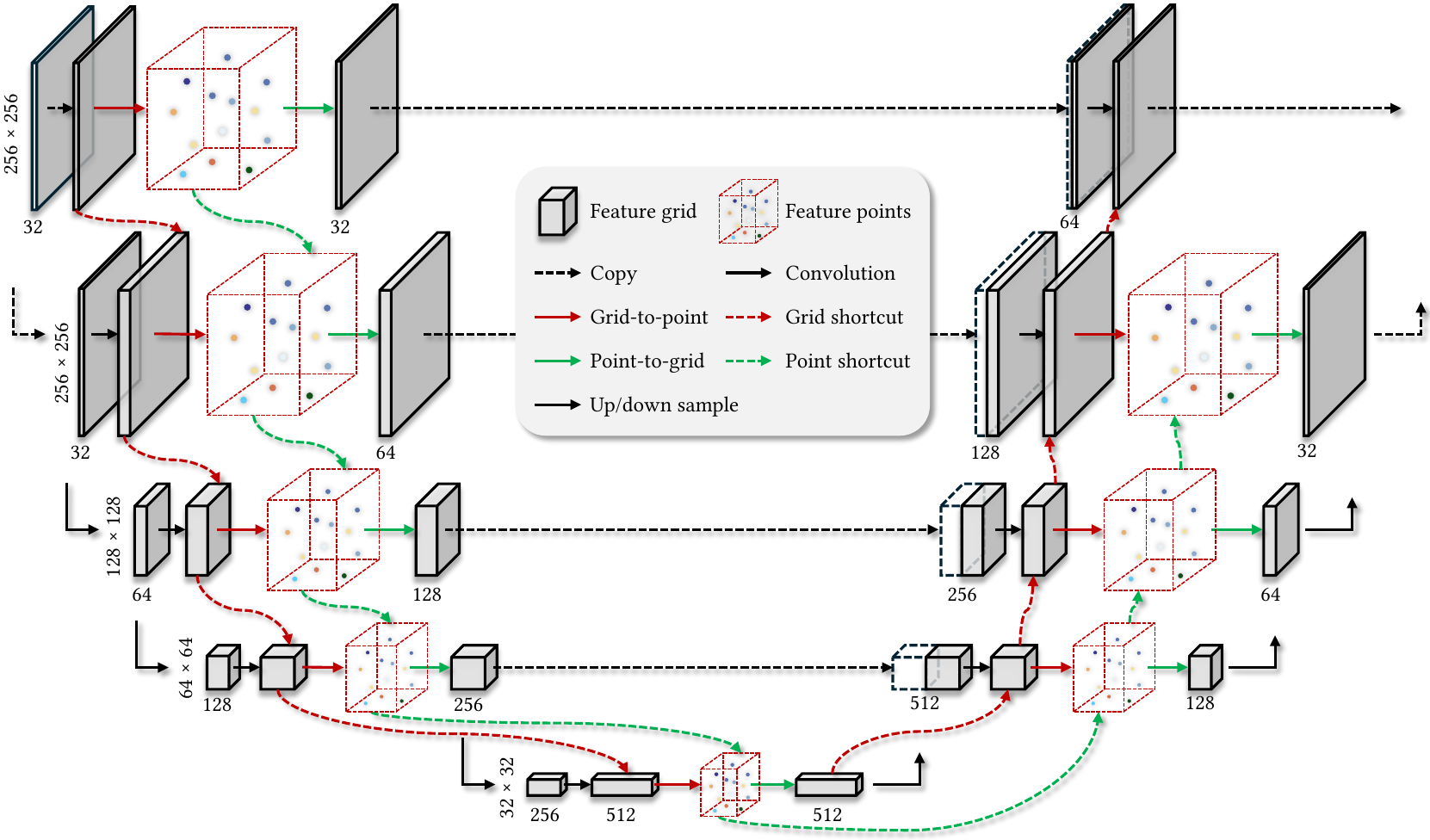,width=0.92\linewidth}}
\caption{Overview of the network architecture for feature reconstruction. The model adopts a multi-scale U-Net-like design with integrated cross-topology refinement. At each scale, grid features and point features exchange information via grid-to-point and point-to-grid transformations (see \autoref{fig:refinement}). Skip connections on both pathways help preserve and propagate fine-grained details across scales.}
\label{fig:alto}
\end{figure*}

\paragraph{Height map decoding}

The refined grid feature is then input to a shallow convolutional network decoder $f_h$ to produce the height map $\hat{\mathbf{H}}$:
\begin{equation}
    \hat{\mathbf{H}} = f_h(\mathbf{G}^{(L)}).
\label{eq:height_decoder}
\end{equation}

\paragraph{Auxiliary decoding}

An auxiliary decoder $f_a$ of another shallow convolutional network is used to predict the building footprint $\hat{\mathbf{A}}$:
\begin{equation}
    \hat{\mathbf{A}} = f_a(\mathbf{G}^{(L)}).
\end{equation}
This auxiliary branch functions to regularize the neural network, promoting more robust predictions when dealing with very noisy and sparse input points.

\subsubsection{Optimization}

We train the model end to end by minimizing the height estimation error, supplemented with an auxiliary footprint loss.

\paragraph{Height reconstruction loss}

We use the mean absolute error between the predicted height map $\hat{\mathbf{H}}$ and the ground truth height map $\mathbf{H}$:
\begin{equation}
    \mathcal{L}_{h} = \frac{1}{M} \sum_{j=1}^{M} \left| \hat{h}_{j} - h_{j} \right|.
\end{equation}

\paragraph{Auxiliary loss}
For building footprint prediction, we apply binary cross-entropy between the predicted footprint probability $\hat{\mathbf{A}}$ and the ground truth $\mathbf{A}$:
\begin{equation}
\begin{split}
\mathcal{L}_{a} = -\frac{1}{M} \sum_{j=1}^{M} [ a_{j} \log \hat{a}_{j} + \, (1 - a_{j}) \log (1 - \hat{a}_{j}) ].
\end{split}
\end{equation}

\paragraph{Total loss}

The total objective is a weighted sum of the height reconstruction loss and the auxiliary loss:
\begin{equation}
    \mathcal{L} = \mathcal{L}_{\mathrm{h}} + \beta \mathcal{L}_{\mathrm{a}},
\label{eq:loss}
\end{equation}
where $\beta$ is a weighting factor.

\paragraph{Post-processing}

\begin{figure}[htb]
  \centering
\centerline{\epsfig{figure=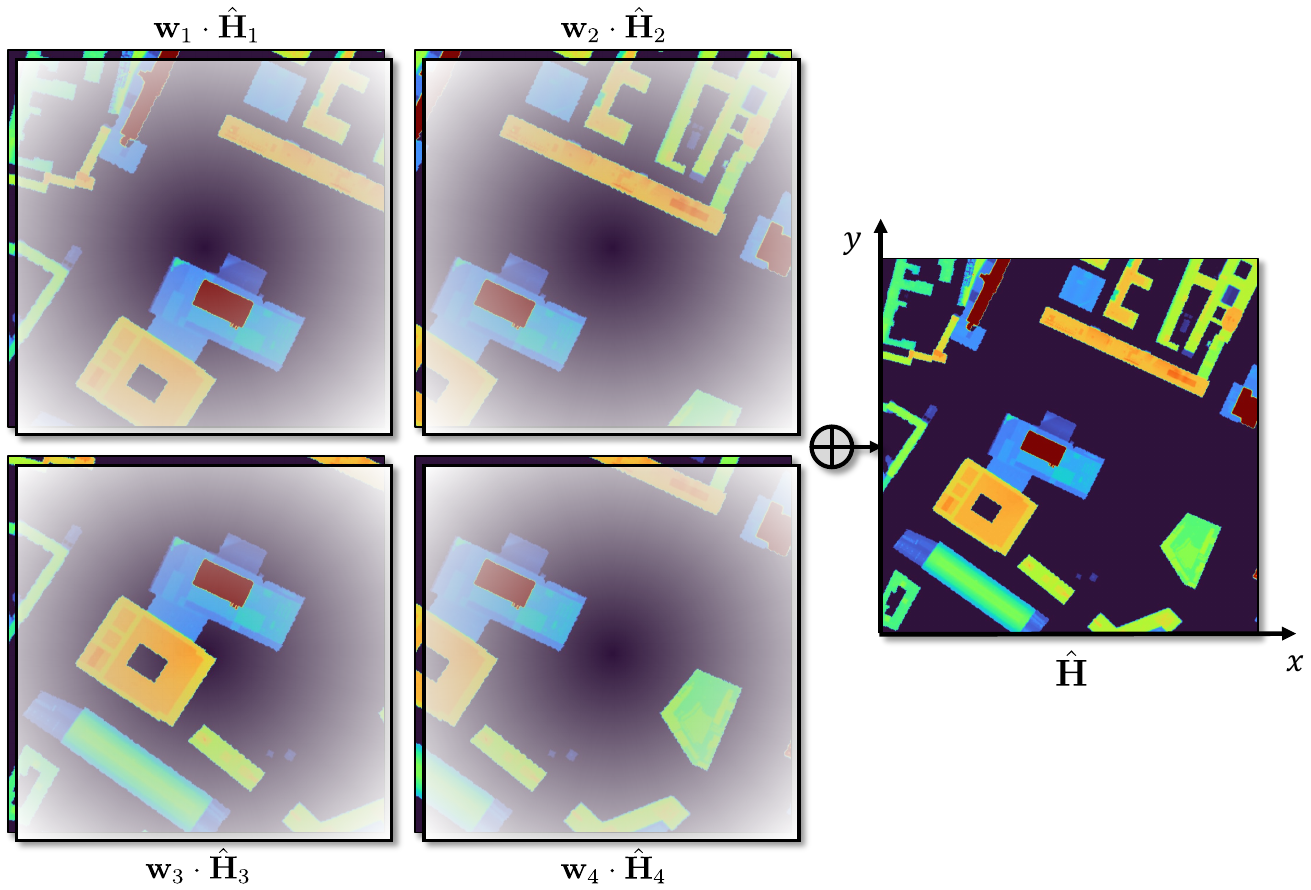,width=0.99\linewidth}}
\caption{Patch blending to form a coherent height map. Multiple overlapping patches, each providing a local height estimate \(\hat{\mathbf{H}}_t\), are weighted by element-wise \(\mathbf{w}_t\) that decrease linearly towards patch edges. These weighted estimates are then averaged to produce a seamless final height map \(\hat{\mathbf{H}}\).}
\label{fig:blending}
\end{figure}

During inference, we predict height maps for overlapping patches of the input region. Each patch provides a local estimate \(\hat{\mathbf{H}}_t\) over its spatial extent. To produce a coherent height map \(\hat{\mathbf{H}}\) without edge artifacts, we mosaic the patches using weighted blending in the overlapping areas, as illustrated in \autoref{fig:blending}. Specifically, let \(\mathbf{w}_t\) denote a spatial blending weight that linearly decreases towards the patch edges. The final height map is then given by:
\begin{equation}
\hat{\mathbf{H}} = \max \left( 0, \frac{\sum {\mathbf{w}_t \cdot \hat{\mathbf{H}}_t}}{\sum \mathbf{w}_t} \right ),
\end{equation}
where $\max(0, \cdot)$ rectifies non-physical predictions since building heights cannot be negative.

\section{Experiments and result analysis}
\label{sec:results}

\subsection{TomoSAR data preprocessing}

\begin{figure}[ht]
  \centering  \centerline{\epsfig{figure=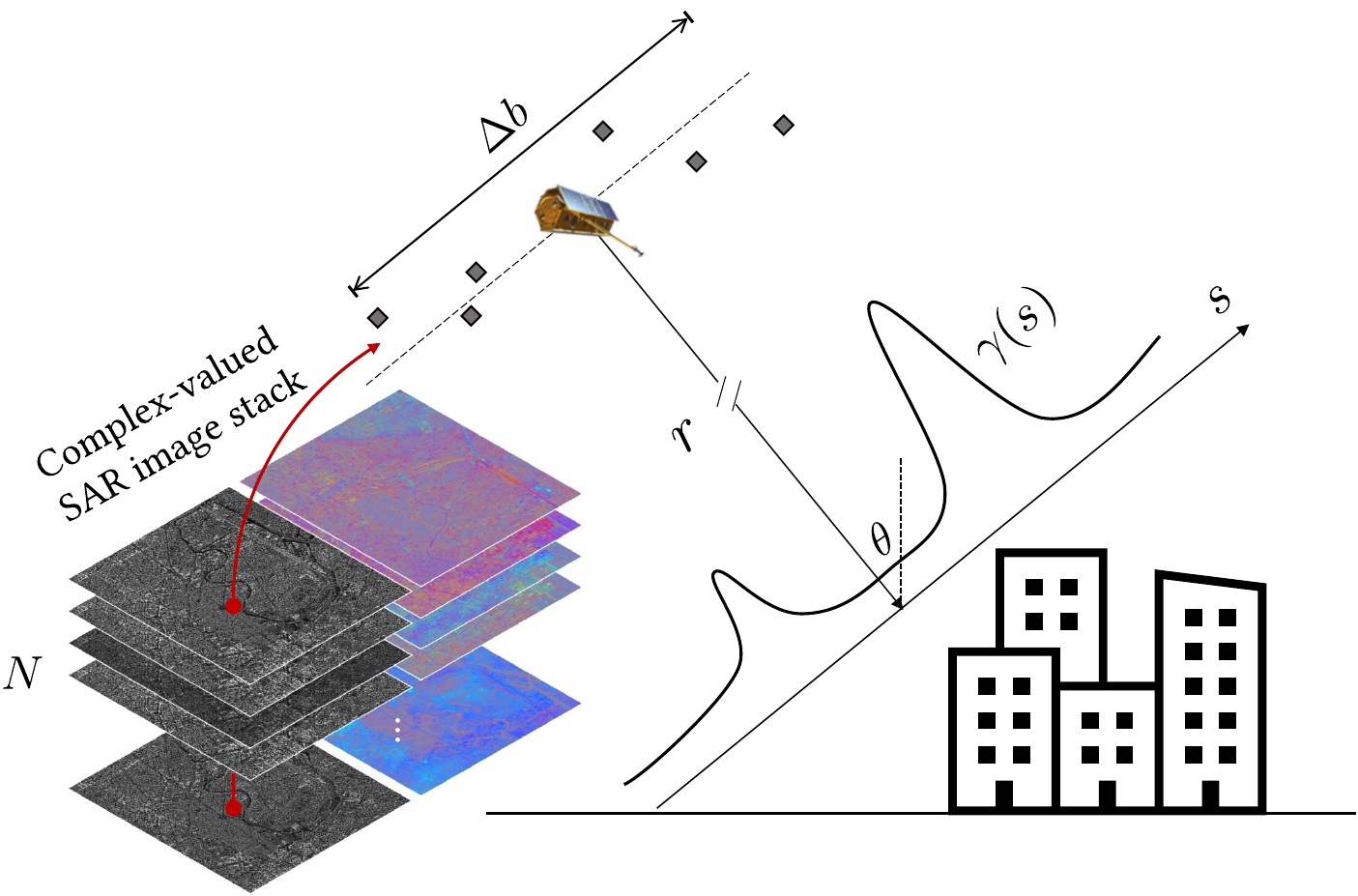,width=0.90\linewidth}}
\caption{Principle of TomoSAR data acquisition and vertical reflectivity reconstruction. Multiple complex-valued SAR images are acquired from different viewing angles, with the maximum elevation aperture \(\Delta b\). TomoSAR reconstructs the reflectivity profile \(\gamma(s)\) along the elevation \(s\) from these images. This process transforms the stack of SAR images into 3D point clouds that characterize building structures. Symbols follow \autoref{tab:stats}.}
\label{fig:tomosar}
\end{figure}

\begin{figure*}[ht]
  \centering  \centerline{\epsfig{figure=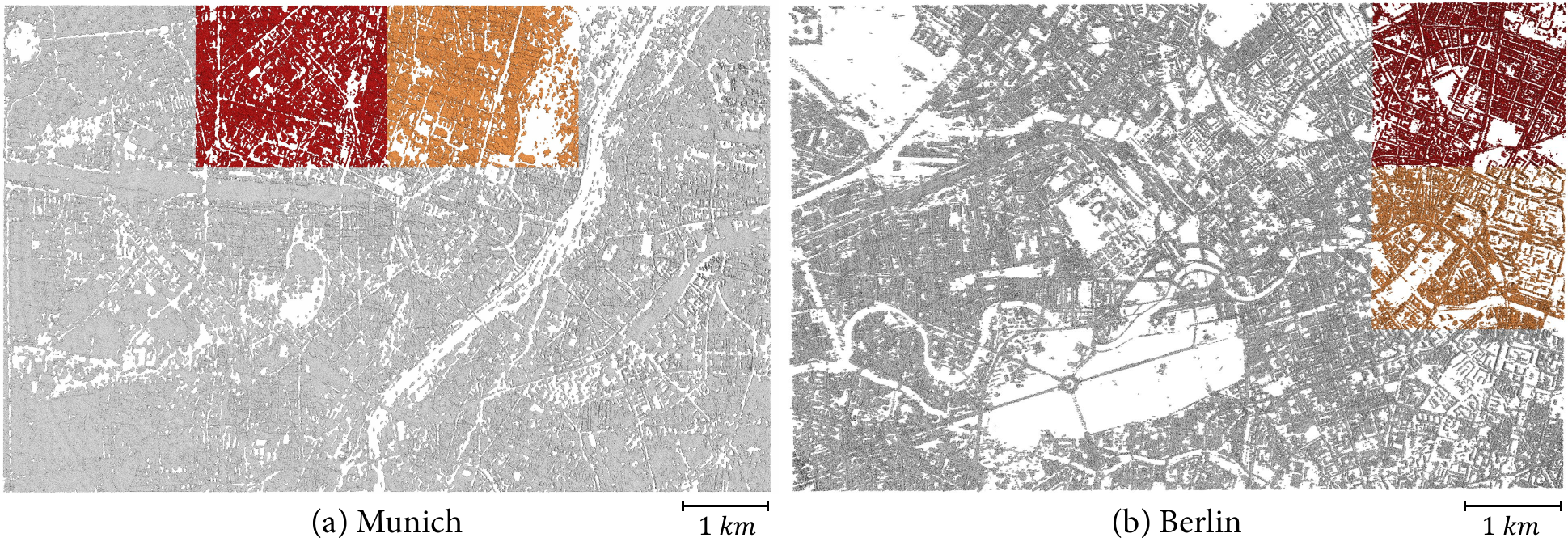,width=0.88\linewidth}}
\caption{Coverage of the TomoSAR point clouds used in this study. Orange (Munich: 4.32\,km$^2$; Berlin: 3.33\,km$^2$) and red regions (Munich: 4.32\,km$^2$; Berlin: 3.25\,km$^2$) are designated for model design validation and final evaluation, respectively, while gray regions are allocated for model training.}
\label{fig:coverage}
\end{figure*}

The SAR data sets consist of a stack of TerraSAR-X high-resolution spotlight images over Berlin, with a spatial resolution of about 1\,m, and a stack of stripmap images over Munich, with a spatial resolution of about 3\,m. The stack of Berlin was acquired between 2008 and 2013 with 108 interferograms, whereas the Munich stack was acquired between 2011 and 2013 and comprises only 5 interferograms. \autoref{tab:stats} summarizes the acquisition parameters, and \autoref{fig:tomosar} illustrates the TomoSAR principle using the same notation. The SAR images were coregistered and corrected for atmospheric phase prior to TomoSAR processing. We perform TomoSAR processing using the ``SVD--Wiener'' algorithm for Berlin \cite{zhu2010very} and the ``NLCS--TomoSAR'' algorithm for Munich \cite{shi_nonlocal}. The processing yields 5D point clouds, consisting of 3D position plus linear deformation rate and seasonal motion amplitude. These two motion components must be accounted for at this resolution to enable precise 3D reconstruction \cite{zhu_timewarp}, as the urban structure and ground surface can undergo displacement due to thermal dilation and subsidence or uplift. More details are provided by Wang \textit{et al.} \cite{wang_fusing_2017} and Shi \textit{et al.} \cite{shi2020sar} for the Berlin and Munich data, respectively. We select downtown areas of Munich and Berlin as study areas due to the availability of complementary data sources, including optical satellite images and nDSM. \autoref{fig:coverage} shows the point clouds for the areas. The data are divided into distinct subsets for training, validation, and testing. To focus exclusively on building heights, we apply cadastral building footprint masks to remove non-building structures from the TomoSAR point clouds and nDSM labels.

\begin{table}[ht]
\centering
\caption{Parameters of SAR data acquisition.}
\label{tab:stats}
\begin{tabular}{lccc}
\toprule
\textbf{Description} & \textbf{Symbol} & \textbf{Munich} & \textbf{Berlin} \\
\midrule
Distance from center & $r$ & 698\,km & 624\,km \\
Wavelength & $\lambda$ & 3.1\,cm & 3.1\,cm \\
Incidence angle & $\theta$ & 50.4$^\circ$ & 36.1$^\circ$ \\
Max elevation aperture & $\Delta b$ & 187\,m & 363\,m \\
Num. of interferograms & $N$ & 5 & 108 \\
\bottomrule
\end{tabular}
\end{table}

\subsection{Experimental setup}

\subsubsection{Implementation details}

The Munich building height nDSM reference was generated using airborne LiDAR data as a reference\footnote{\url{https://geodaten.bayern.de/opengeodata/}}, while the Berlin data were obtained from official photogrammetric sources\footnote{\url{https://gdi.berlin.de/}}. Both have a spatial resolution of 1\,m. We train the model using the Adam optimizer with weight decay. The loss weight $\beta$ in \autoref{eq:loss} is fixed at 10 in all experiments. During training, we sample patches on the fly for a total of 10{,}000 steps by drawing random centers within the corresponding region and rejecting any sample whose window would cross a region boundary. This ensures that no patch spans two splits and prevents leakage between train, validation, and test. For efficiency, each region is stored as several non-overlapping chunks on disk, and each patch is sampled within a single chunk, keeping the additional overhead negligible. We select the checkpoint with the best validation performance and evaluate it on the test set. We use a cyclic learning-rate schedule with 1{,}000-step cycles, halving the maximum learning rate after each cycle.

\subsubsection{Evaluation metrics}

To comprehensively assess the reconstruction performance of our model, we use the following three error metrics:

\begin{itemize}
    \item Mean absolute error (MAE): This metric measures the average magnitude of the errors in a set of predictions, without considering their direction. It is calculated as:
    \begin{equation}
        \text{MAE} = \frac{1}{M} \sum_{j=1}^{M} \left| h_j - \hat{h}_j \right|.
    \end{equation}
    \item Root mean square error (RMSE): RMSE provides a more sensitive measure to larger errors by squaring the differences before averaging and taking the square root. It is calculated as:
    \begin{equation}
        \text{RMSE} = \sqrt{ \frac{1}{M} \sum_{j=1}^{M} \left( h_j - \hat{h}_j \right)^2 }.
    \end{equation}
    \item Median absolute error (MedAE): MedAE focuses on the median of the absolute errors, being more robust to outliers. It is calculated as:
    \begin{equation}
        \text{MedAE} = \operatorname{median}_{j=1}^{M} \left| h_j - \hat{h}_j \right|.
    \end{equation}
\end{itemize}
We report these metrics across the following regions to provide nuanced performance indicators:
\begin{itemize}
    \item Overall area: These are computed over all pixels in the test split. The overall metrics are reported by default unless stated otherwise.

    \item Building area: These are computed over pixels within building footprints.

    \item Building instance: These are computed per building by taking the median predicted height within each footprint, and then aggregated across instances.

\end{itemize}
\noindent Here, \( h_j \) and \( \hat{h}_j \) denote the actual and the predicted height value at pixel \( j \), and \( M \) is the total number of evaluated pixels.

\begin{table*}[ht]
\centering
\caption{Quantitative height reconstruction results for Berlin and Munich (m).}
\label{tab:error_analysis}
\begin{tabular}{lccccccccc}
\toprule
\multirow{2}{*}{\textbf{City}} & \multicolumn{3}{c}{\textbf{Overall Area}} & \multicolumn{3}{c}{\textbf{Building Area}} & \multicolumn{3}{c}{\textbf{Building Instance}} \\
\cmidrule(lr){2-4} \cmidrule(lr){5-7} \cmidrule(lr){8-10}
 & \textbf{MAE} & \textbf{RMSE} & \textbf{MedAE} & \textbf{MAE} & \textbf{RMSE} & \textbf{MedAE} & \textbf{MAE} & \textbf{RMSE} & \textbf{MedAE} \\
\midrule
\textbf{Berlin} & 2.10 & 5.46 & 0.00 & 4.64 & 8.12 & 1.55 & 3.69 & 6.17 & 2.32 \\
\textbf{Munich} & 3.27 & 6.44 & 0.04 & 6.38 & 9.04 & 4.26 & 5.06 & 6.87 & 3.31 \\
\bottomrule
\end{tabular}
\label{tab:result_main}
\end{table*}

\subsection{Reconstruction from Berlin data}

\begin{figure*}[ht]
  \centering  \centerline{\epsfig{figure=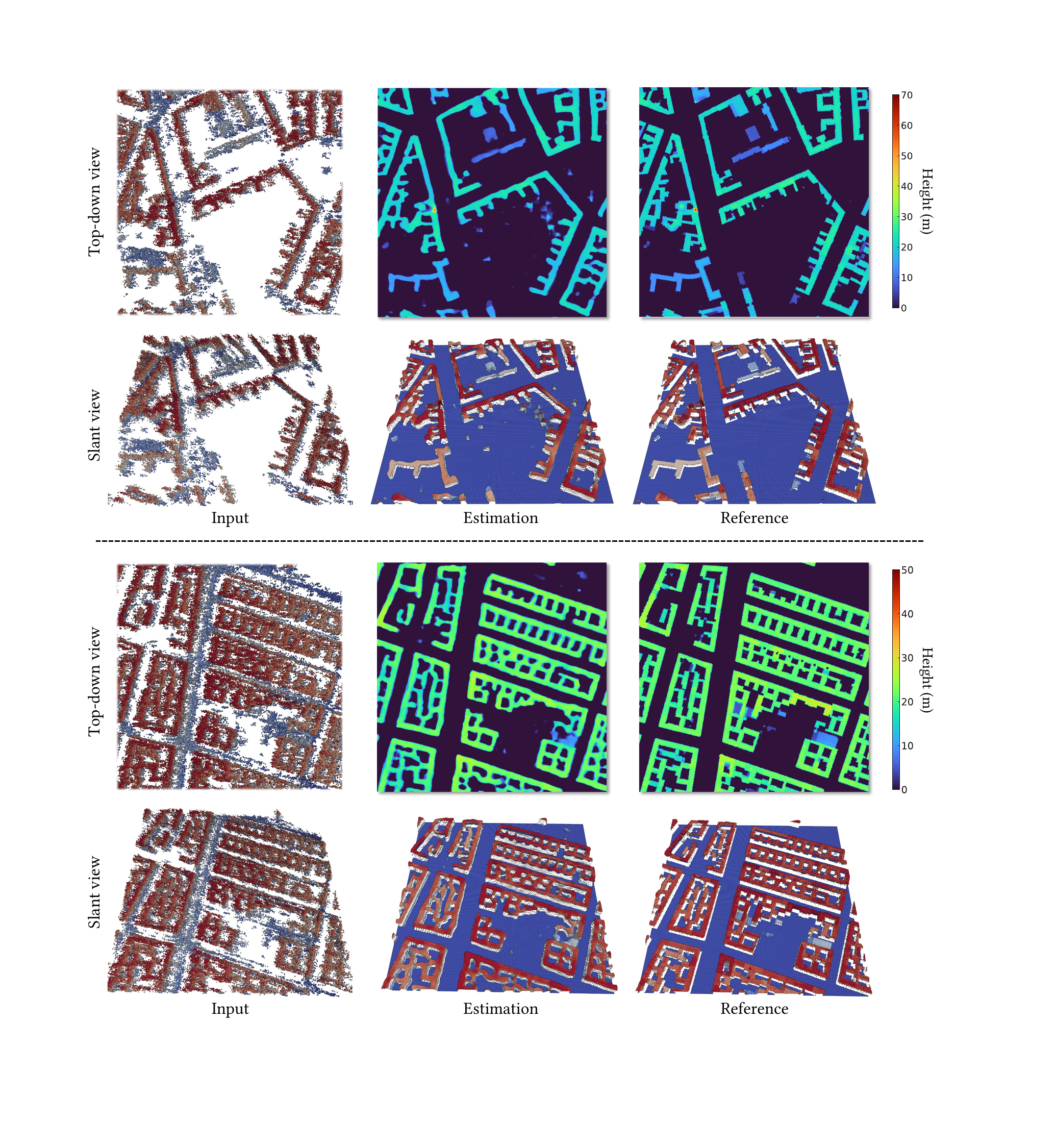,width=0.82\linewidth}}
\caption{Qualitative height reconstruction results for two areas in Berlin. In the slant views, the height maps are reprojected as 3D points. The results demonstrate the model's robustness to noisy inputs and anisotropic sampling.}
\label{fig:result_berlin}
\end{figure*}

\autoref{tab:result_main} reports the errors over the test area, with an overall MAE of 2.10\,m achieved on the Berlin dataset. \autoref{fig:result_berlin} presents qualitative results for two selected patches, while \autoref{fig:result_overview} provides an overview of the results for the entire test area. These results demonstrate that the model effectively inpaints missing regions in the input point cloud, recovering most structures accurately. However, the error distribution in \autoref{fig:distribution} shows that the model exhibits higher error rates when processing more complex building structures.

\subsection{Reconstruction from Munich data}

\begin{figure*}[ht]
  \centering  \centerline{\epsfig{figure=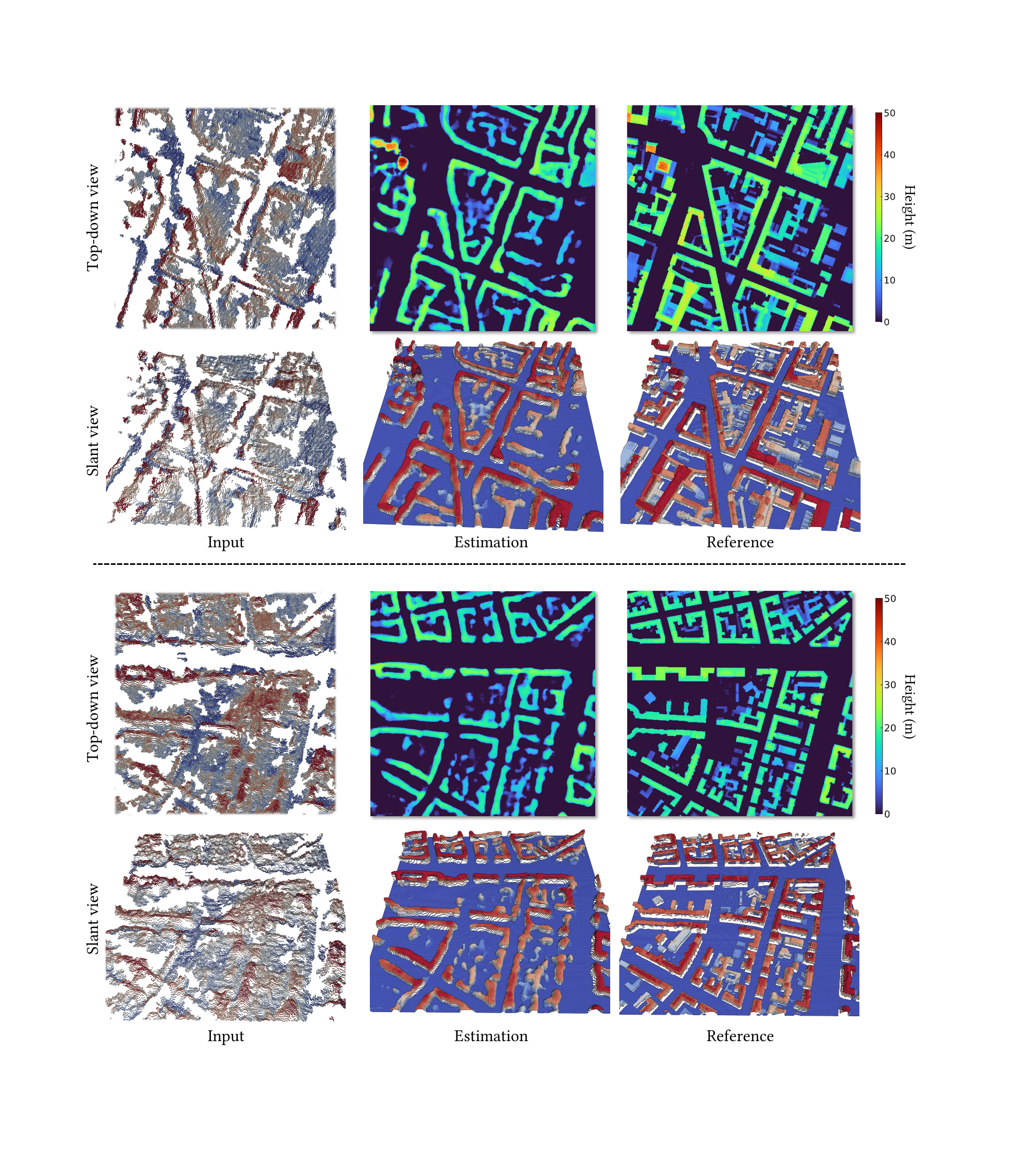,width=0.82\linewidth}}
\caption{Qualitative height reconstruction results for Munich. In the slant views, the rasterized results are reprojected as 3D points. The results demonstrate the model's robustness in handling noisy inputs with anisotropic spatial distributions.}
\label{fig:result_munich}
\end{figure*}

A straightforward approach would be to apply the same model to the Munich data. However, due to the more severe noise, sparsity, and the highly anisotropic distribution of the Munich points, the baseline method that predicts only a height map fails to deliver reasonable results, as shown in \autoref{tab:ablation_component}. A closer inspection reveals that the predicted heights, before non-negative rectification, cluster around zero, indicating that the network struggles to identify meaningful building signals. We address this issue by adding auxiliary supervision from a building footprint mask, which regularizes training by encouraging the model to distinguish samples inside and outside building footprints. These footprint masks are used only during training; at inference the model relies exclusively on raw TomoSAR points, enabling deployment at scale even when auxiliary data are unavailable.

\begin{figure*}[ht]
  \centering
\centerline{\epsfig{figure=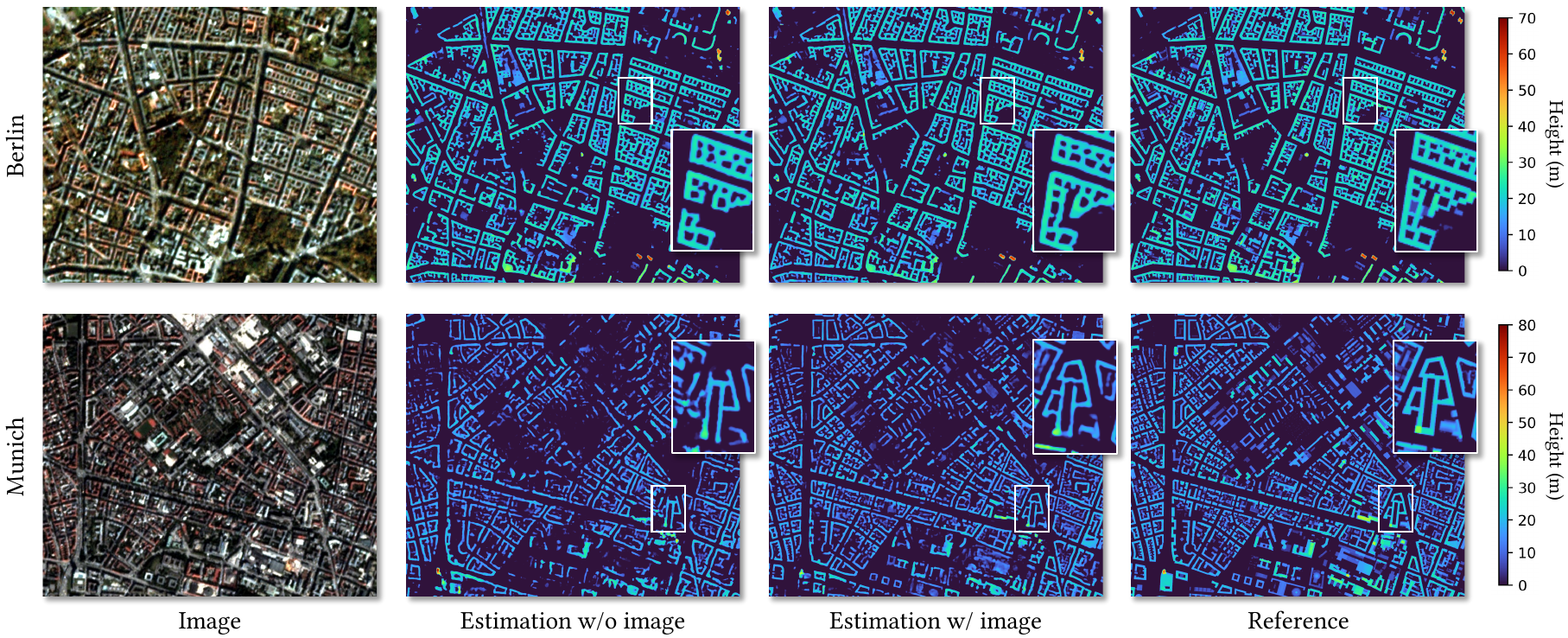,width=0.96\linewidth}}
\caption{Reconstruction results over the complete test areas of both datasets. Incorporating optical image features improves accuracy, particularly for fine details and in regions with sparse or uneven point coverage. Quantitative results are reported in \autoref{tab:result_rgb}.}
\label{fig:result_overview}
\end{figure*}

As shown in \autoref{tab:result_main}, the reconstructed height map for Munich is of lower fidelity than that for Berlin, with an MAE of 3.27\,m. \autoref{fig:distribution} further indicates that errors increase with building height, underscoring the challenge of modeling taller structures. In \autoref{fig:result_munich} and \autoref{fig:result_overview}, it is evident that the boundaries are less regularized and some built-up areas are missing due to the absence of input points. Nonetheless, the predicted height map reveals clearer urban structures that are largely obscured by noise in the raw point clouds. Considering the quality of the input data, these results suggest that the approach can exploit even challenging TomoSAR stacks. Compared with Berlin, Munich's stripmap SAR data represent a more cost-effective data acquisition method that could enable broader coverage and large-scale mapping products.

\subsection{Ablation study}

\begin{figure}[ht]
  \centering  \centerline{\epsfig{figure=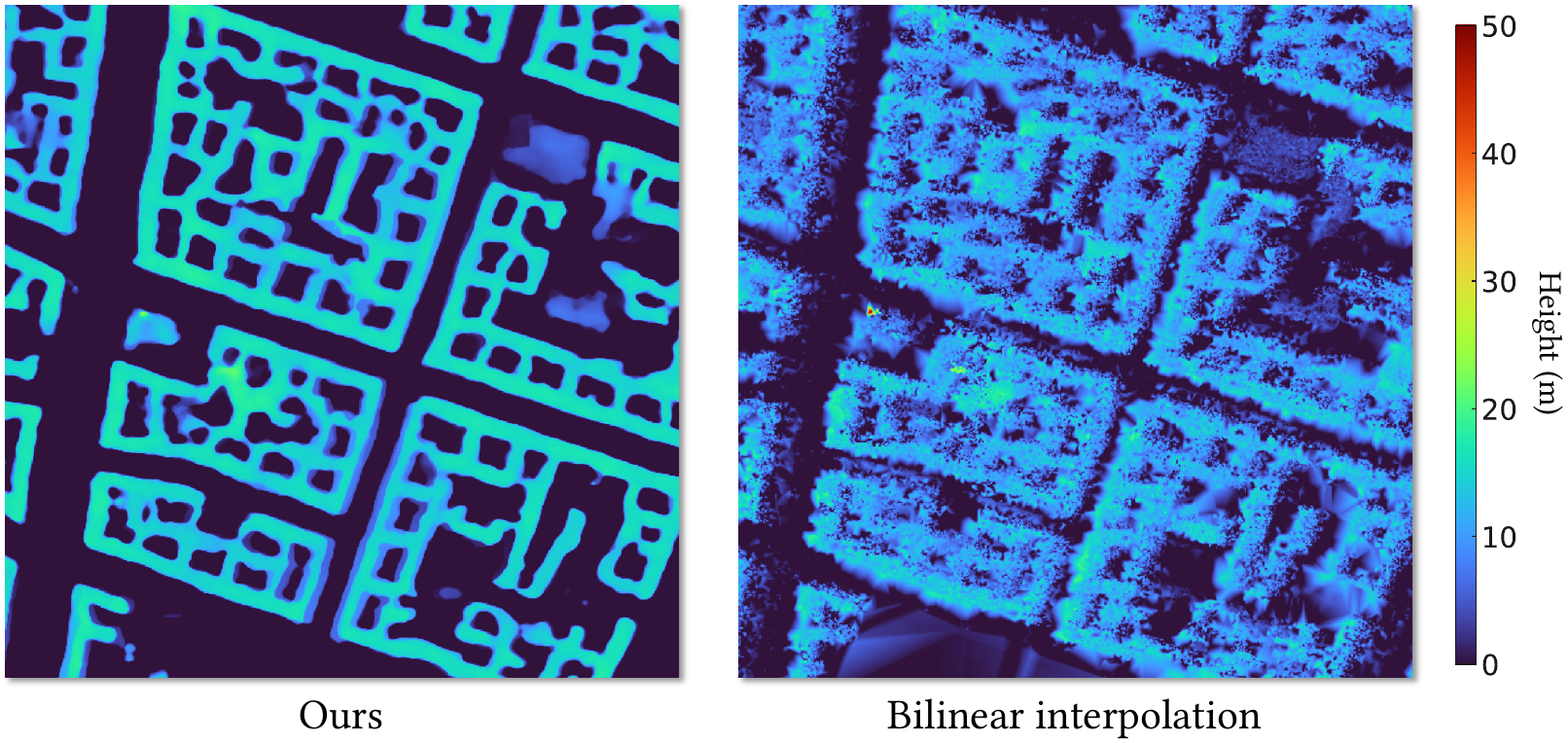,width=0.99\linewidth}}
\caption{Comparison of height maps from our method and bilinear interpolation at a Berlin site. Bilinear interpolation relies on a DTM, whereas we use raw points at inference, yielding a more accurate reconstruction.}
\label{fig:interpolation}
\end{figure}

\begin{table}[ht]
\centering
\caption{Evaluation of interpolation methods versus our approach. Bilinear interpolation and inverse distance weighting struggle under significant noise and anisotropic data distribution and require a DTM at inference time. Our network achieves lower error while using the DTM only during training data preparation.}
\label{tab:interpolation}
\begin{tabular}{lcc|cc}
\toprule
\multirow{2}{*}{\textbf{Method}} & \multicolumn{2}{c|}{\textbf{DTM required}} & \multicolumn{2}{c}{\textbf{MAE} (m)} \\
\cmidrule(lr){2-3}\cmidrule(lr){4-5}
 & \textbf{Train} & \textbf{Infer} & \textbf{Berlin} & \textbf{Munich} \\
\midrule
\cellcolor{lightblue}\textbf{Our neural network} & \cellcolor{lightblue}\checkmark & \cellcolor{lightblue}\xmark & \cellcolor{lightblue}\textbf{2.10} & \cellcolor{lightblue}\textbf{3.27} \\
\textbf{Bilinear interpolation} & n/a & \checkmark & 5.44 & 6.84 \\
\textbf{IDW interpolation}      & n/a & \checkmark & 5.53 & 6.85 \\
\bottomrule
\end{tabular}
\end{table}

\autoref{tab:interpolation} compares our approach with interpolation methods. Both bilinear interpolation and inverse distance weighting perform poorly, as they fail to capture the noise and anisotropic point distribution of the data. It can be seen from \autoref{fig:interpolation} that bilinear interpolation produces very noisy height values. Moreover, these explicit techniques depend on DTM to mitigate terrain effects. In contrast, our proposed neural network learns a robust, data-driven mapping that can directly produce building height maps without requiring the DTM as input, making it a more versatile solution.

\begin{figure*}[ht]
  \centering  \centerline{\epsfig{figure=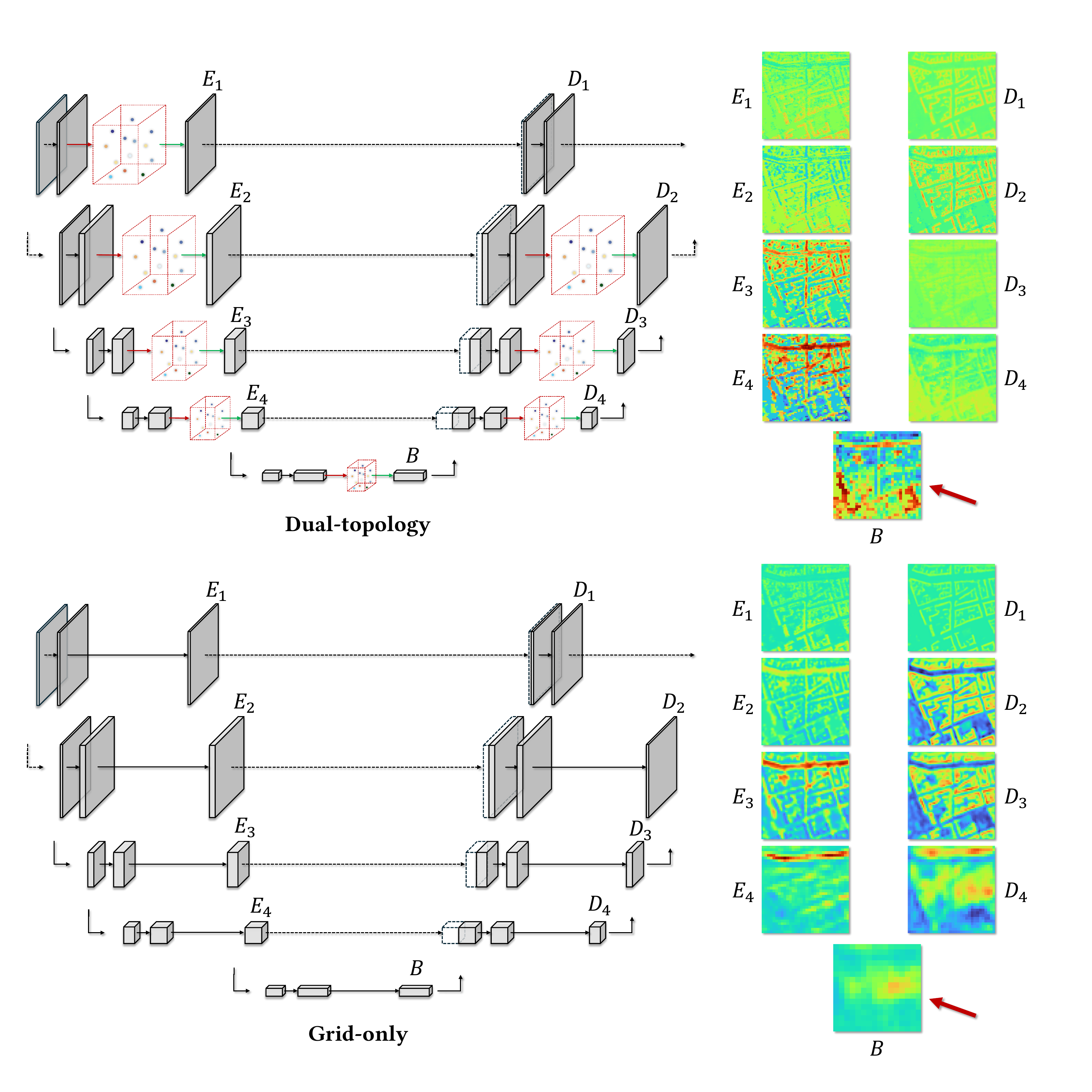,width=0.78\linewidth}}
\caption{Comparison of intermediate feature maps between the dual-topology network and a grid-only U-Net. Each column pair \((E_i, D_i)\) shows the encoder and decoder feature maps at a given scale, while \(B\) denotes the bottleneck feature map. With cross-topology refinement, the dual-topology network preserves high-frequency features exhibiting clearer building structures and more distinct spatial patterns.}
\label{fig:featuremap}
\end{figure*}

\autoref{fig:featuremap} compares the intermediate feature maps from our dual-topology network and a vanilla U-Net based solely on grid topology. With cross-topology refinement, the dual-topology network preserves sharper structural cues and more distinct spatial patterns. \autoref{tab:ablation_component} further demonstrates the impact of point-grid transformations and the auxiliary footprint branch. On both datasets, incorporating the point topology improves performance. Interestingly, the supervision from the mask prediction branch is effective only on the Munich data, likely because only the lower-quality data benefits from additional regularization. In contrast, the Berlin data does not benefit in the same way; adding the auxiliary task can shift the learning objective away from height prediction. Since the performance difference is marginal, this option can be left to the user based on data quality. Notably, the PointNet encoder with 2D local pooling~\cite{peng2020eccv} outperforms the more resource-intensive PointNet++ variant~\cite{qi2017pointnet++} that uses 3D aggregation, suggesting that leveraging local structural context on the regular grid is particularly beneficial for height estimation.

\autoref{tab:ablation_config} summarizes the impact of network depth (\textit{i.e.}, $L$ as in \autoref{eq:height_decoder}) and feature plane resolution. For Berlin, a 5-layer network is sufficient to capture the essential structure. Increasing depth does not improve performance but substantially increases parameter count and training cost. In contrast, for Munich, the lower-quality data benefits from increased depth, indicating a need for stronger context modeling. Balancing accuracy and complexity, we set the network depth to 6 layers. Additionally, feature map resolution also affects performance: higher resolution can better preserve feature details, it also requires stronger inpainting to compensate for missing values, whereas lower resolution requires less effort for the neural network to complete the missing values but can smooth out structure. The two factors counteract each other and therefore there is a balancing point. For both Berlin and Munich data, the resolution of $256\times256$ performed the best.

\begin{table}[ht]
\centering
\caption{Ablation of neural network components. Light blue marks our default configurations. ``Aux.'' denotes auxiliary footprint supervision.}
\label{tab:ablation_component}
\begin{tabular}{cccc|c}
\toprule
\multicolumn{3}{c}{\textbf{Component}} & \multicolumn{2}{c}{\textbf{MAE} (m)} \\
\cmidrule(r){1-3} \cmidrule(l){4-5}
\textbf{Aux.}  & \textbf{Point Topology} & \textbf{Encoder} & \textbf{Berlin} & \textbf{Munich} \\
\midrule
\cellcolor{lightblue}\checkmark & \cellcolor{lightblue}\checkmark & PointNet w/ local pool & 2.16 & \textbf{3.27} \\
\cellcolor{lightblue}\xmark & \cellcolor{lightblue}\checkmark  & PointNet w/ local pool & \textbf{2.10} & 4.76 \\
\checkmark & \xmark  & PointNet w/ local pool & 2.43 & 3.40 \\
\xmark & \xmark  & PointNet w/ local pool & 2.38 & 4.76 \\
- & \checkmark  & PointNet++ & 3.47 & 4.24 \\
\bottomrule
\end{tabular}
\end{table}

\begin{table}[ht]
\centering
\caption{Ablation of network depth and feature-plane resolution. Light blue marks our default configurations.}
\label{tab:ablation_config}
\begin{tabular}{lccc|cc}
\toprule
\multirow{2}{*}{\textbf{Configuration}} & \multirow{2}{*}{\textbf{Value}} & \multirow{2}{*}{\textbf{\#Params}} & \multicolumn{2}{c}{\textbf{MAE} (m)} \\
\cmidrule(lr){4-5}
 &  &  & \textbf{Berlin} & \textbf{Munich} \\
\midrule
\multirow{3}{*}{\textbf{Depth}} 
    & \cellcolor{lightblue}5 & \cellcolor{lightblue}11.1\,M & \textbf{2.10} & 3.31 \\
    & \cellcolor{lightblue}6 & \cellcolor{lightblue}43.4\,M & 2.13 & 3.27 \\
    & 7 & 172.3\,M & 2.27 & \textbf{3.25} \\
\midrule
\multirow{3}{*}{\textbf{Resolution}} 
    & 128 & 43.4\,M & 2.28 & 3.34 \\
    & \cellcolor{lightblue}256 & \cellcolor{lightblue}43.4\,M & \textbf{2.10} & \textbf{3.27} \\
    & 512 & 43.4\,M & 2.18 & 3.30 \\
\bottomrule
\end{tabular}
\end{table}

\subsection{Incorporating optical satellite images}

\begin{figure}[ht]
  \centering  \centerline{\epsfig{figure=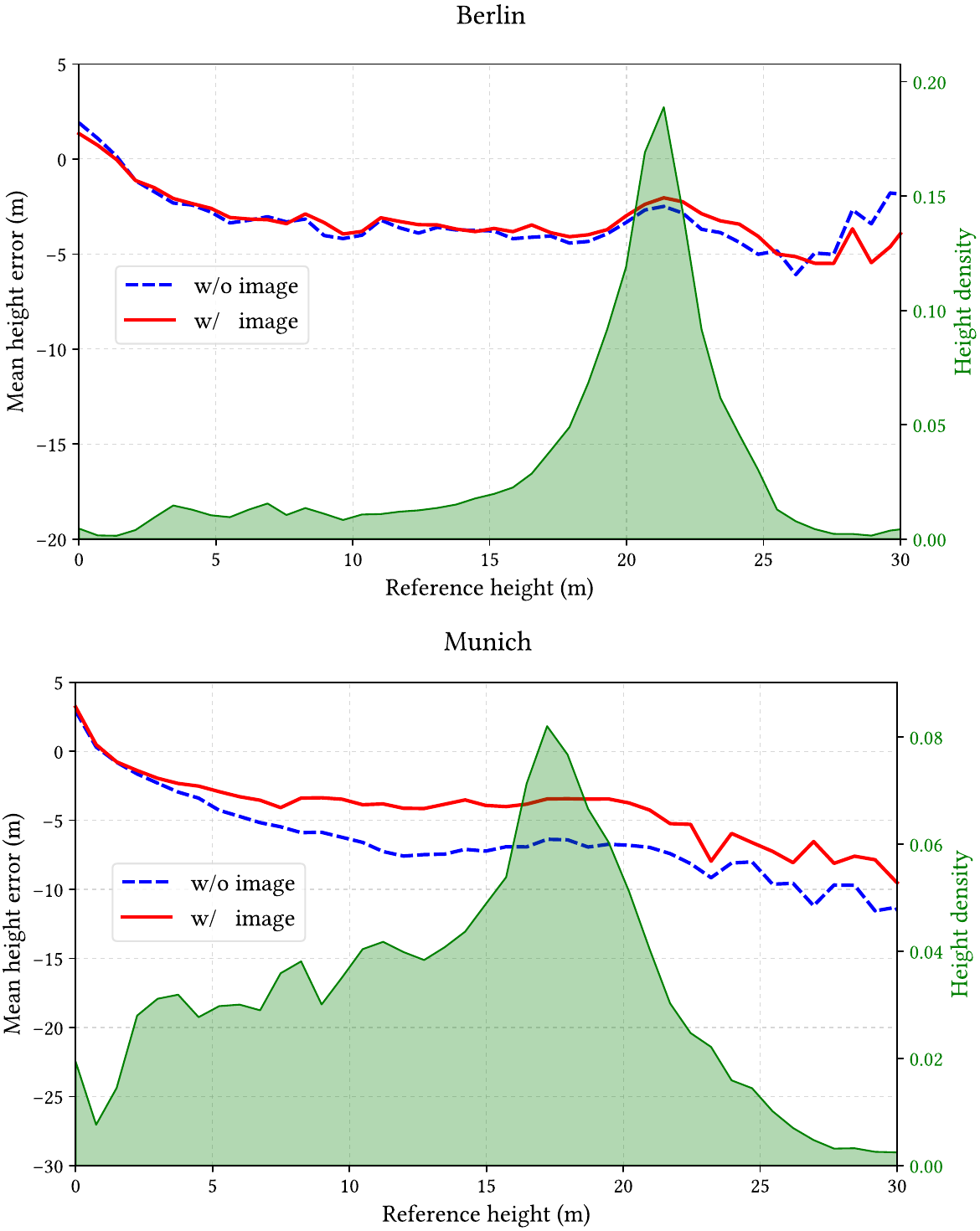,width=0.99\linewidth}}
\caption{Reconstruction errors with respect to ground truth height values. Incorporating semantic information from images reduces errors, especially for the Munich data. For clarity, only height values within the 0--30\,m range are shown, with zero-height values excluded.}
\label{fig:distribution}
\end{figure}

Our framework is extensible and allows optical imagery to be integrated into the pipeline. To demonstrate this capability, we incorporate PlanetScope optical satellite images with a resolution of 3--5\,m~\cite{team2017planet} whose earliest scenes in our study area date back to 2017. Like TomoSAR point clouds, these images can provide large-scale coverage. We encode the images with a 6-layer U-Net to produce grid-aligned features that match the dimensionality of the TomoSAR grid features, as depicted in \autoref{fig:architecture}. \autoref{tab:result_rgb} reports the results when using imagery as an additional input. The geometric information from TomoSAR point clouds and the semantic information from imagery complement each other, yielding improved height predictions when both sources are used compared with using either source alone. As shown in \autoref{fig:result_overview}, fusing both sources produces more regularized predictions. The results also highlight the benefit of integrating TomoSAR point clouds into image-based pipelines. \autoref{fig:distribution} further demonstrates that adding the images leads to lower reconstruction errors across the full height range, with greater gains on the lower-quality Munich dataset. The imagery branch is backbone-agnostic and remains optional (\textit{e.g.}, in cases of cloud cover), and this design enables us to exploit all available data without compromising large-scale deployment. The flexibility of our framework comes from projecting features onto a nadir grid, which makes such fusion straightforward. This experiment is intended to demonstrate extensibility rather than to optimize performance.

\begin{table}[ht]
\centering
\caption{Mean absolute error (m) for different input sources. TomoSAR point geometry ($\mathbf{P}$) and image-based semantics ($\mathbf{I}$) are complementary; combining them yields the lowest errors.}
\label{tab:result_rgb}
\begin{tabular}{lcc|cc|cc}
\toprule
\multirow{2}{*}{\textbf{Input}} & \multicolumn{2}{c}{\textbf{Overall Area}} & \multicolumn{2}{c}{\textbf{Building Area}} & \multicolumn{2}{c}{\textbf{Building Instance}} \\
\cmidrule(lr){2-3} \cmidrule(lr){4-5} \cmidrule(lr){6-7}
 & \textbf{Berlin} & \textbf{Munich} & \textbf{Berlin} & \textbf{Munich} & \textbf{Berlin} & \textbf{Munich} \\
\midrule
\multirow{1}{*}{$\mathbf{P}$} 
    & 2.10 & 3.27 & 4.64 & 6.38 & 3.69 & 5.06 \\
\midrule
\multirow{1}{*}{$\mathbf{I}$} 
    & 2.37 & 2.54 & 5.31 & 5.12 & 4.61 & 3.46 \\
\midrule
\multirow{1}{*}{\textbf{$\mathbf{P \& I}$}} 
    & \textbf{2.00} & \textbf{2.18} & \textbf{4.46} & \textbf{4.54} & \textbf{3.54} & \textbf{3.31} \\
\bottomrule
\end{tabular}
\end{table}

\section{Discussions}

\subsection{Technical justifications and limitations}

In this section, we discuss several key technical aspects and limitations of our proposed method.

\paragraph{Inductive biases} Our method benefits from the inductive biases of CNNs, which encourage spatially coherent representations even when TomoSAR points are noisy and anisotropically sampled. Although the design is conceptually straightforward, the model learns robust structural priors that enable reliable height reconstruction under challenging conditions, including the heavily degraded Munich stripmap data. We also find that the network provides a degree of translation invariance: the model remains stable under co-registration errors of up to several pixels (equivalently, a few meters) in the input points. This behavior further suggests that the learned priors help the network recover plausible height maps despite imperfect alignment. While our validation currently covers only two cities, we expect the approach to generalize to other urban areas with similar SAR stack characteristics (\textit{e.g.}, number of interferograms, imaging mode, baseline distribution), as all point clouds are derived from the same imaging modes (spotlight or stripmap) using physics-based processing.

\paragraph{Data fusion} The network can integrate multiple input data sources as long as their features can be mapped onto the orthogonal plane. In our experiments, adding PlanetScope optical imagery provides complementary semantic cues and improves accuracy, especially on the lower-quality Munich data where more sparse and noisy points leave more room for image-guided refinement. This unified representation makes it straightforward to fuse 2D and 3D inputs. Nevertheless, modality-specific issues remain, such as cloud cover in optical images, and overall reconstruction quality is still bounded by the fidelity of the input data.

\paragraph{Label consistency} To limit temporal changes, we focus the evaluation on the most stable urban regions. We also filter out all non-building elements from the nDSM, including terrain and other infrastructure such as bridges and roads. This preprocessing can introduce noise into the reference height map, compounded by disparities between data sources, which may contribute to the systematic underestimation observed in \autoref{fig:distribution}. More broadly, the effectiveness of our pipeline still depends on the accuracy and consistency of the reference height map. It is worth noting that the same pipeline could be applied to include other stationary objects, provided that corresponding labels are available. However, this would introduce additional challenges in separating objects like trees or cars in the reference. \autoref{tab:type} breaks down MAE by building type. The modest and directionally mixed gaps indicate that performance variations are driven by structural complexity and local data quality rather than by building type.

\begin{table}[ht]
\centering
\caption{Mean absolute error (m) by building type.}
\label{tab:type}
\begin{tabular}{cc|cc}
\toprule
\multicolumn{2}{c}{\textbf{Berlin}} & \multicolumn{2}{c}{\textbf{Munich}} \\
\cmidrule(r){1-2} \cmidrule(l){3-4}
\textbf{Residential} & \textbf{Non-residential} & \textbf{Residential} & \textbf{Non-residential} \\
\midrule
 4.67       & 4.55          & 6.09       & 6.55            \\
\bottomrule
\end{tabular}
\end{table}

\subsection{Future opportunities}

Motivated by the aforementioned limitations, we outline two directions for future work.

\paragraph{Uncertainty-aware modeling} TomoSAR point clouds are typically derived via model-based inversion, where the uncertainty of the estimates is well formulated and can be quantified from the input data quality (\textit{e.g.}, SNR). Beyond the strongly anisotropic errors, noise levels also vary substantially across points due to the large dynamic range of SAR observations. This heteroscedastic uncertainty is not yet explored in our network design. At present, we treat all input points equally. A natural extension is to incorporate per-point uncertainty as an additional input attribute and to use it to reweight samples during training. Moreover, predicting uncertainty alongside height could further improve robustness, as suggested by recent uncertainty-aware learning approaches, while also providing an interpretable confidence measure for downstream use.

\paragraph{Scalable multi-sensor integration} This study focuses on 2.5D building representations in the form of height maps. However, compared with LiDAR, TomoSAR can capture much more detailed building facade information, which could be utilized for full-scale 3D reconstruction, with height estimation as only one component of its broader potential. In combination with other Earth observation data sources, multi-scale representations of the built environment warrant further exploration. Promising directions include fusing TomoSAR point clouds with LiDAR or photogrammetric point clouds for complete 3D modeling, and combining optical imagery with TomoSAR for object-level reconstruction. Meanwhile, high-quality nDSM data are not always available for supervision. In such cases, weakly or self-supervised strategies may help sustain performance. The former can leverage lower-quality or proxy elevation sources, while the latter can exploit intrinsic constraints such as geometric consistency. Finally, footprint supervision is only one possible auxiliary cue for regularizing training; additional cues remain to be explored. Together, these directions could enable learning in label-scarce regions and improve scalability beyond well-mapped areas.

\section{Conclusion}

We have presented a framework for reconstructing building height maps from spaceborne TomoSAR point clouds using a dual-topology network design over point and grid representations. Experiments on two TomoSAR datasets of varying quality show that our approach effectively denoises the input points and inpaints missing values to produce high-fidelity height maps. Moreover, the framework is readily extensible to incorporate satellite optical imagery, which provides complementary cues and further improves reconstruction quality. As a proof of concept, our method demonstrates strong potential to advance large-scale building height mapping.

\bibliographystyle{IEEEtran}
\bibliography{refs}

\end{document}